\pdfoutput=1

\documentclass[]{article}
\usepackage{multirow}
\usepackage[normalem]{ulem}
\usepackage{tabularray}
\usepackage{graphicx}
\usepackage{subcaption}
\usepackage{color}
\usepackage{comment}
\usepackage{acl}
\usepackage{amsmath, amsthm, amssymb, paralist}
\usepackage{times}
\usepackage{latexsym}
\usepackage{tcolorbox}
\usepackage[linesnumbered,ruled,vlined]{algorithm2e}
\usepackage{float} 
\usepackage[T1]{fontenc}

\usepackage[utf8]{inputenc}
\usepackage{inconsolata}
\usepackage{microtype}
\usepackage{graphicx}
\usepackage{comment}
\usepackage{multirow}
\usepackage{arydshln}
\usepackage{booktabs}
\usepackage{colortbl}
\usepackage{color}
\definecolor{Iron}{rgb}{0.851, 0.855, 0.863}%

\title{Interactive Memory Learning for Long-Term Conversations}

\author{
    Cai Ke$^{1, 2}$, 
    Jiangyue Yan$^{1,2}$,
    Han Zhang$^{2}$, 
    Xin Liu$^{2}$\footnotemark[1], 
    Zike Yuan$^{1,2}$,
    \textbf{Yue Yu}$^2$\textbf{,}\\
    \textbf{Hui Wang}$^2$\textbf{,}  
     \textbf{and Ruifeng Xu}$^{1,2}${\hypersetup{hidelinks}\thanks{\quad Corresponding authors.}} \\
    $^1$Harbin Institute of Technology, Shenzhen, China \quad $^2$Pengcheng Laboratory, China \\
    \texttt{kecai@stu.hit.edu.cn, xuruifeng@hit.edu.cn}
}

\begin{document}
\maketitle

\begin{abstract}

Recent advancements in large language models have significantly enhanced the capabilities of agents in modeling long-term conversations. Despite these successes, existing approaches typically adopt a static heuristic paradigm, where information is passively archived without adaptive memory valuation. Consequently, these methods fail to self-evolve or align their memory management with evolving user needs. To address this, we propose \textbf{\textsc{Icml}} (\textbf{I}ntera\textbf{C}tive \textbf{M}emory \textbf{L}earning), a multi-agent framework that transforms the memory mechanism from a passive archive into a learnable, interactive memory policy. Specifically, we first employ a session synthesis pipeline to generate expert data, facilitating rapid test-time adaptation in unseen scenarios. Building on this, \textbf{\textsc{Icml}} utilizes an online reinforcement learning mechanism where a \texttt{Planner} agent selectively encodes high-value information and a \texttt{Trigger} agent dynamically retrieves it to optimize response quality, whereby the two agents co-evolve through continuous interaction feedback. Crucially, both agents are synchronized through a delayed reward mechanism that propagates future feedback back to earlier storage decisions, ensuring memory policies are precisely aligned with user expectations. Experimental results demonstrate that \textbf{\textsc{Icml}} significantly outperforms strong baselines, exhibiting the unique capability to continuously improve response quality as interactions accumulate.

\end{abstract}

\section{Introduction}

The remarkable advances in Large Language Models (LLMs) have led to the rapid development of open-domain conversations~\citep{li2017dailydialog,zhang2018personalizing,dinan2018wizard,rashkin2019towards,baumgartner2020pushshift}. By modeling historical information, LLMs demonstrate strong capabilities in generating fluent responses. However, in long-term conversations, they still struggle to maintain human-like engagement. A major reason is the lack of an effective mechanism that allows the model to truly learn and adapt through continuous dialogue~\citep{xu2022beyond,shi2023large,zhang2024bench,du-etal-2024-perltqa,li2024long,levy-etal-2024-task,liu2024lost,zhang2025survey,hu2025memory}. 

\begin{figure}[!t]
\centering
\includegraphics[width=\linewidth]{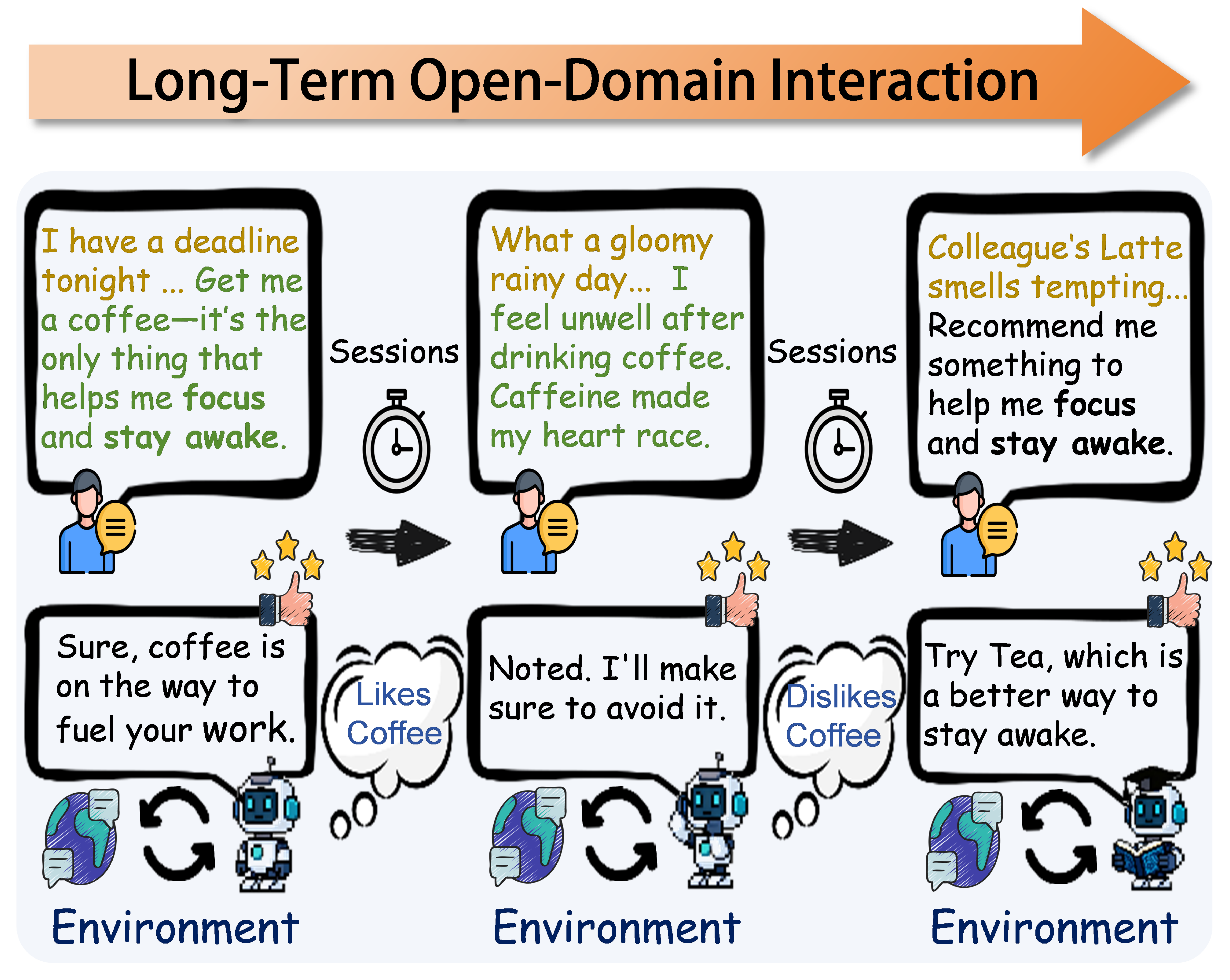}
\caption{The \textbf{\textsc{Icml}} framework for long-term conversations. Through long-term interactions, the agent utilizes environmental feedback to distinguish \textcolor[RGB]{88,142,49}{high-value memories} from \textcolor[RGB]{181,139,2}{low-value noise} for online self-evolution.}%
\label{intro}
\end{figure}

The essence of long-term open-domain conversation generation is the ability to satisfy the user's constantly changing expectations and preferences over time. This requires a dynamic memory process where the agent learns from real-time interactions to provide personalized services. Most existing methods, however, rely on a static heuristic paradigm~\cite{bae2022keep,jang2023conversation,zhang2023mind,lu2023memochat,zhong2024memorybank,li-etal-2025-hello,ong2024towards,chen2025compress,wang2025recursively,ke2025flexibly,ke2026dynamic,liang2026meta}. These methods treat memory as a simple database rather than a learning process, failing to understand what to memorize and when to trigger. This leads to unresolved conflicts between outdated preferences and new user requirements.

In contrast, Cognitive Psychology suggests that human memory is not a passive archive but a learnable process. Humans do not treat all information equally; instead, we selectively encode information that has high value for future decisions while discarding irrelevant noise~~\citep{schank1980language,Tulving1983-TULEOE,tulving2002episodic,anderson2005cognitive,yadav2022prefrontal}. Furthermore, we continuously update our memory through feedback to adapt to changing circumstances. As illustrated in Figure~\ref{intro}, humans naturally distinguish between \textcolor[RGB]{88,142,49}{high-value memories} (e.g., a critical health warning like heart race) and \textcolor[RGB]{181,139,2}{low-value noise} (e.g., transient states like a deadline or gloomy rainy day). Consequently, when the user's situation changes, the listener actively updates their mental model by letting the new constraint override the outdated preference. This self-evolution capability allows humans to become more understanding as interactions progress. \textbf{Therefore, we argue that the key to mastering long-term conversations lies in transforming passive heuristic paradigm into an interactive memory learning paradigm where the agent learns what to memorize and when to trigger based on continuous environmental feedback}.

To realize this goal, we introduce \textbf{\textsc{Icml}} (\textbf{I}ntera\textbf{C}tive \textbf{M}emory \textbf{L}earning), a novel multi-agent collaborative framework underpinned by online Reinforcement Learning (RL). Specifically, addressing the challenge where agents typically lack relevant memories when encountering a new environment for the first time interaction and thus produce suboptimal responses, we devise a retrospective session synthesis pipeline. Starting from a seed session involving user initial interaction as an expert demonstration, we inversely generate multiple consistent storylines comprising interconnected sessions, which are subsequently forward-annotated to produce high-quality expert data for autonomous test-time adaptation. The core of \textbf{\textsc{Icml}} consists of two interactive Actor-Critic~\citep{konda1999actor} agents: a \texttt{Planner} agent that selectively memorizes high-value information, and a \texttt{Trigger} agent that retrieves memory based on the utterance. Crucially, these agents co-evolve to align memory planning with actual utility through a delayed cross-session truth reward mechanism. While the \texttt{Planner} makes initial storage decisions, its policy is refined only when the \texttt{Trigger} successfully utilizes the memory to satisfy user expectations.
This feedback loop ensures that both agents mutually adapt and converge toward an optimal collaborative strategy for personalized engagement. Experimental results on three long-term conversation datasets derived from real human interactions demonstrate that \textbf{\textsc{Icml}} significantly outperforms strong baselines, exhibiting the capability to effectively evolve into a more personalized agent over time. \textbf{The contribution can be summarized as follows:}

1) We explore a learnable memory paradigm that leverages continuous environmental feedback to dynamically optimize what to memorize and when to trigger.%

2) We are the first to propose a plug-and-play, online RL framework for long-term open-domain conversation that enables autonomous test-time adaptation of memory policies, allowing the model to self-evolve and align with user expectations without human intervention.

3) Extensive evaluations on three long-term open-domain datasets demonstrate that \textsc{Icml} significantly outperforms state-of-the-art baselines, with response quality and personalization improving consistently as the agent evolves through continuous interaction.

\begin{figure*}[!t]
\centering
\includegraphics[width=\textwidth]{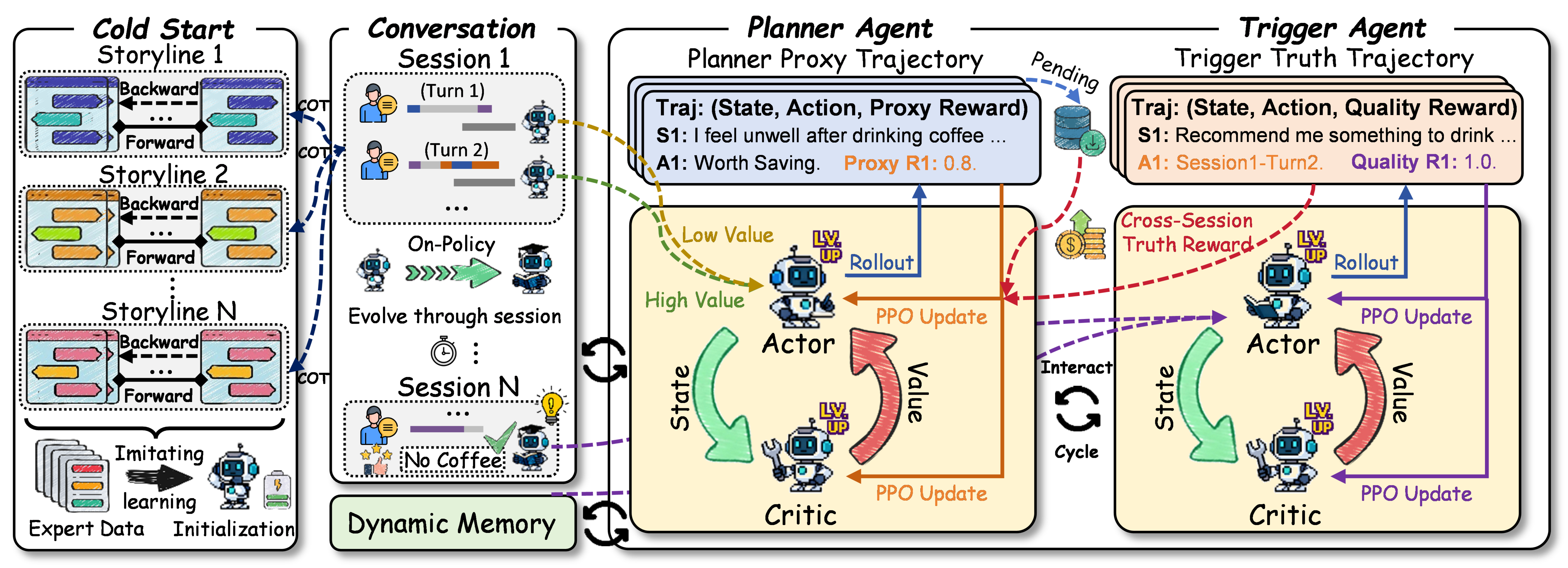}
\caption{Illustration of our \textbf{Retrospective Session Synthesis} (Left) and \textbf{\textsc{Icml}} framework (Right).}
\label{method}
\end{figure*}

\section{Related Work}
\paragraph{Long-Term Open-Domain Conversation.}
Long-term open-domain conversation generation~\cite{xu2022beyond,jang2023conversation,zhang2023mind} aims to simulate real-world human-to-human interactions, focusing on building lifelong companionship and personalized experiences rather than long-term question answering. To achieve this, a major trend is developing generation-centric dialogue agents~\cite{lu2023memochat,zhong2024memorybank,chen2025compress,li-etal-2025-hello,ong2024towards,wang2025recursively,anonymous2026thinkflow} for LLMs. For example, existing methods often compress dialogue sessions into static summaries or specific user facts \citep{zhong2024memorybank, li-etal-2025-hello}. Moreover, some methods also explore recursive summarization \citep{wang2025recursively} or model the impact of time lines~\citep{zhang2023mind,ong2024towards} to maintain consistency over time. \textbf{Different from these methods relying on passive storage and retrieval, we propose a new paradigm to selectively encode and retrieve high-value information via environmental feedback, thereby achieving self-evolution through online RL}.
\paragraph{Agentic Memory Architectures and Management.}
Prior works on memory management have explored various mechanisms for management-centric memory agents~\cite{packer2023memgpt,liu2024agentlite,mei2024aios,wang2025mem,chhikara2025mem0,xu2025mem,kang-etal-2025-memory}, focusing on designing sophisticated architectures to handle the full lifecycle of memory. For instance, \citet{chhikara2025mem0} utilize graph-based representations to capture complex relational structures. \citet{xu2025mem} link memories as structured notes that dynamically evolve through interconnected indexing. Moreover, \citet{kang-etal-2025-memory} introduce an OS-inspired hierarchical storage system comprising short-, mid-, and long-term units. \textbf{Different from these heuristic architectures, we introduce a collaborative multi-agent framework where the agent teams co-evolve via a delayed reward mechanism, ensuring memory policies are precisely aligned with user expectations in long-term conversations}.

\section{Methodology}
\label{sec:methodology}

We approach the long-term conversation task as a sequential decision-making problem, where the agent must learn to dynamically manage its memory to maximize long-term conversational quality. Our method, as shown in Figure~\ref{method}, consists of three key components: (1) \textbf{Problem Formulation}, which rigorously defines the interactive memory learning process as a Partially Observable Markov Decision Process (POMDP)~\citep{aastrom1965optimal}; (2) \textbf{Retrospective Session Synthesis}, a novel data synthesis pipeline for initializing the system with high-quality expert data to address the cold-start problem; and (3) The \textbf{\textsc{Icml}} framework, comprising collaborative \texttt{Planner} and \texttt{Trigger} agents that continuously evolve via \textbf{Cross-Session Truth Rewards}.

\subsection{Problem Formulation}
\label{subsec:problem_formulation}

To rigorously model the dynamic interaction where the agent must infer user intent from limited context and memory, we formulate the \textbf{Interactive Memory Learning} process as a POMDP, defined by the tuple $\langle \mathcal{S}, \mathcal{A}, \mathcal{O}, \mathcal{R}, \gamma \rangle$, where $\mathcal{R}$ denotes the reward function and $\gamma$ represents the discount factor.

\paragraph{State and Observation.}
The underlying state $s_k \in \mathcal{S}$ at turn $k$ includes the user's latent intent and the complete interaction history, which is not fully visible to the agent. Instead, the agent receives an observation $o_k \in \mathcal{O}$, consisting of the current user query $u_k$, the recent dialogue context $H_k$, and the current external memory state $\mathcal{M}_k = \{m_1, m_2, \dots, m_N\}$.

\paragraph{Action Space.}
The joint action $a_k=(a^p_k, a^t_k)$ decomposes into a memory planning action $a^p_k \in \{0, 1\}$ and a memory triggering action $a^t_k \in \{0, 1, \dots, N\}$ controlled by the \texttt{Planner} and \texttt{Trigger} respectively.

\subsection{Retrospective Session Synthesis}
\label{subsec:backward_construction}

Training a robust POMDP policy requires high-quality data where memory dependencies are explicit (i.e., knowing \textit{why} a memory was saved and \textit{when} it was used). However, test-time interaction datasets typically lack long-term consistency labels, and initializing the policy from scratch often leads to the cold-start problem, where the agent fails to capture critical user constraints due to sparse rewards. To overcome this, we introduce a \textbf{Retrospective Session Synthesis} pipeline to synthesize expert data with dense, causal memory dependencies by utilizing the chain of thought~\cite{wei2022chain}. All prompts are shown in Appendix~\ref{data_prompt}.

\paragraph{Backward Storyline Generation.}
To address the challenge where agents lack relevant memories during their initial interaction in a new environment, we adopt a reverse-generation strategy. We designate the initial interaction where a new user first reveals specific constraints or preferences as the \textbf{seed session} $S_{seed}$. Using backbone LLMs as user simulators, we then recursively generate preceding sessions $S_{prev}$ that logically ground the context of $S_{seed}$ (e.g., generating a past event where coffee caused heart race). This reverse causality ensures that the generated history $\mathbb{H}_{gen} = \{S_{prev}^{(L)}, \dots, S_{prev}^{(1)}, S_{seed}\}$ maintains strict logical consistency, providing high-quality expert data that explain the origin of current user preferences.

\paragraph{Forward Dependency Annotation.}
With the coherent storyline established, we traverse $\mathbb{H}_{gen}$ in chronological order to generate ground-truth labels for supervised warm-up.

\begin{itemize}
    \item For the \texttt{Planner}, we evaluate the information gain of each turn to assign a binary label $y^p \in \{0, 1\}$, indicating whether the turn contains high-value information worth saving.
    \item For the \texttt{Trigger}, we identify the specific historical fragments required to resolve the query in $S_{seed}$, assigning the target retrieval index $y^t$. To enhance the \texttt{Trigger}'s discrimination ability against semantic noise, we further mix this ground-truth with hard negatives (irrelevant turns from the same session) and soft negatives (random global memories).
\end{itemize}

This pipeline yields a high-quality expert dataset $\mathcal{D}_{expert}$, which is utilized to initialize the policy $\pi_\theta$ before online deployment.

\subsection{The \textsc{Icml} Framework}
\label{subsec:icml_framework}

As illustrated in Figure~\ref{method}\footnote{Note that in the trajectory tuples, $State$ refers to the agent's observation state representation.}, the core of \textbf{\textsc{Icml}} consists of two collaborative agents—the \texttt{Planner} and the \texttt{Trigger}, which co-evolve to align memory management with actual conversational utility. This architecture enables autonomous test-time adaptation without human intervention, allowing the model to refine its policies during live interactions. To stabilize the online learning process in this complex interactive environment, we adopt a shared Actor-Critic~\citep{konda1999actor} that governs two collaborative agents. All prompts and pseudocode are shown in Appendix~\ref{training_prompt} and~\ref{algorithm}.

\subsubsection{The Planner Agent (Memory Planning)}
\label{subsubsec:planner}

The \texttt{Planner} acts as the proactive gatekeeper of long-term memory. Its primary goal is to identify and retain high-value memories while filtering out low-value noise.

\paragraph{State and Policy.}
At turn $k$, the \texttt{Planner} receives an observation $o^p_k$ consisting of the user query $u_k$.%
The policy $\pi_{\theta}(a^p_k | o^p_k)$ outputs a binary distribution over action space $\{0, 1\}$:

\begin{itemize}
    \item \textbf{Save ($a^p_k=1$):} The current interaction is condensed into a memory fragment $m_{new}$ and appended to the external memory.
    \item \textbf{Discard ($a^p_k=0$):} The information is deemed redundant or irrelevant and is discarded.
\end{itemize}

\paragraph{Proxy Reward.}
Since the true utility of a memory is often unknown at the moment of storage, we employ backbone LLMs to provide an immediate \textit{proxy reward} $r^{proxy}_k \in [0, 1]$. This judge evaluates the intrinsic information value of the turn, providing a dense signal to guide the \texttt{Planner}'s exploration in the early stages. Additionally, we incorporate a miss-penalty term: if the \texttt{Planner} discards a high-value turn, a negative reward $-\alpha r^{proxy}_k$ is applied to discourage information loss during exploration.

\subsubsection{The Trigger Agent (Memory Triggering)}
\label{subsubsec:trigger}

The \texttt{Trigger} is responsible for contextualizing the generation process by retrieving the most relevant information from the dynamic memory. Unlike traditional dense retrieval, the \texttt{Trigger} learns a policy to select memories that maximize the final response quality.

\paragraph{State and Policy.}
The \texttt{Trigger} observes the current query $u_k$, the dialogue history $H_k$, and a set of candidate memories $\mathcal{M}_k$. The policy $\pi_{\theta}(a^t_k | o^t_k)$ outputs a categorical distribution over the memory indices $\{0, 1, \dots, |\mathcal{M}_k|\}$. Selecting index $0$ implies no memory is needed. The selected memory $m_{a^t_k}$ is then concatenated with the context to generate the final response.

\paragraph{Quality Reward.}
To accurately evaluate the agent's performance, we do not rely on simple heuristics. Instead, we also employ backbone LLMs to score the final response based on multiple dimensions, yielding a comprehensive \textit{quality reward} $r^{qual}_k \in [0, 1]$. This multi-dimensional scoring aligns the \texttt{Trigger}'s objective with complex human preferences.

\subsubsection{Response Generation}
Finally, our \textbf{\textsc{Icml}} generates a personalized response $r^*$ by grounding the LLM in the retrieved memory $m_{a^t_k}$ and current context:
\begin{equation}
    r^* \sim P_{\text{LLM}}( \cdot \mid H_k, u_k, m_{a^t_k} ).
\end{equation}
This process bridges temporal gaps across sessions and yields $r^{qual}$, serving as the ultimate feedback to drive the co-evolution of the entire system.

\subsubsection{Cross-Session Truth Reward}
\label{subsubsec:truth_reward}

To resolve the delayed verification of memory utility, we propose the \textbf{Cross-Session Truth Reward} mechanism. It propagates the quality signal from $r^*$ back to the \texttt{Planner}'s historical storage decisions, aligning memory policies with actual utility. We maintain a \textit{Pending Reward Buffer} that stores the \texttt{Planner}'s latent experiences (i.e., stored memories waiting to be verified). 

When the \texttt{Trigger} activates a memory fragment $m_i$ at a future turn $k_{future}$ to address a user query, we retrospectively trace $m_i$ back to its creation turn $k_{past}$. We then propagate the obtained quality assessment $r^{qual}_{k_{future}}$ back to the \texttt{Planner} as the \textit{truth reward}:
\begin{equation}
    r^{truth}_{k_{past}} = r^{proxy}_{k_{past}} + \lambda \cdot r^{qual}_{k_{future}} \cdot \mathbb{I}(m_i \text{ is triggered}),
\end{equation}
where $\lambda$ is a weighting factor. This mechanism aligns the \texttt{Planner}'s storage objective with the long-term utility of the memory. By linking historical planning with future retrieval success, the \texttt{Planner} and \texttt{Trigger} mutually adapt their policies, ensuring the internal memory state is precisely aligned with latent user expectations.

\subsubsection{On-Policy Optimization}
\label{subsubsec:optimization}

We employ the Proximal Policy Optimization (PPO) algorithm~\citep{schulman2017proximal} for end-to-end optimization. During the online interaction phase, the agent performs rollouts through in real-world scenarios, collecting experience trajectories $\tau = \{ (o_k, a_k, r_k) \}_{k=1}^{T}$.

The optimization objective involves maximizing the cumulative return $R^j_k = \sum_{i=k}^{T} \gamma^{t-k} r^j_i$, where $j \in \{p, t\}$ denotes the \texttt{Planner} or \texttt{Trigger} agent. The Critic loss $L_{critic}(\phi)$ minimizes the mean squared error between the estimated value $V_\phi^j(o^j_k)$ and the actual return:
\begin{equation}
L_{critic}(\phi) = \sum_{j \in \{p, t\}} \mathbb{E}_{\tau} \left[ \sum_{k=0}^{T} \left( V_\phi^j(o^j_k) - R^j_k \right)^2 \right].
\end{equation}

The Actor loss $L_{actor}(\theta)$ is computed over the collected trajectories using the clipped surrogate objective:
\begin{equation}
\begin{aligned}
 L_{actor}(\theta) = \sum_{j \in \{p, t\}} & \mathbb{E}_{(o,a) \sim \tau} \Big[ \min \big( \rho^j_k A^j_k, \\
 & \text{clip}(\rho^j_k, 1-\epsilon, 1+\epsilon) A^j_k \big) \\
 & + \beta \mathbb{S}[\pi^j_\theta](o^j_k) \Big],
\end{aligned}
\end{equation}
where $\rho^j_k = \frac{\pi^j_\theta(a^j_k|o^j_k)}{\pi^j_{\theta_{old}}(a^j_k|o^j_k)}$ is the importance sampling ratio, $A^j_k$ is the advantage estimated based on $R^j_k$, and $\mathbb{S}$ denotes the entropy bonus. This joint optimization allows both agents to co-evolve their specific policies.

\begin{table*}[!t]
\renewcommand{\arraystretch}{1.2}
\small
\centering
\setlength{\tabcolsep}{1.1mm}{
\begin{tabular}{cccccc|cccc|cccc} 
\specialrule{1.5pt}{0pt}{0pt}
\hline
\multirow{2}{*}{\textbf{Backbone}}       & \multirow{2}{*}{\textbf{Methods}} & \multicolumn{4}{c}{\textbf{CC}}                                  & \multicolumn{4}{c}{\textbf{MSC}}                                 & \multicolumn{4}{c}{\textbf{GC}}                                   \\ 
\cline{3-14}
                                         &                                   & \textbf{B-4}  & \textbf{R-L}   & \textbf{Bert}  & \textbf{Mauve} & \textbf{B-4}  & \textbf{R-L}   & \textbf{Bert}  & \textbf{Mauve} & \textbf{B-4}  & \textbf{R-L}   & \textbf{Bert}  & \textbf{Mauve}  \\ 
\hline
\multirow{19}{*}{\textbf{GPT-4o}}        & Long Context (128K)                & 1.79          & 17.41          & \textbf{47.79} & 55.73          & 1.21          & 15.12          & \textbf{49.17} & 54.36          & 0.66          & \textbf{11.43} & 36.57          & 25.12           \\ 
\cdashline{2-14}
                                          & Mem0~\citeyearpar{chhikara2025mem0}                              & 1.02          & 14.57          & 45.85          & 46.92          & 0.69          & 12.78          & 45.68          & 45.61          & 0.53          & 9.42           & 34.08          & 23.43           \\
                                         & A-Mem~\citeyearpar{xu2025mem}                             & 1.21          & 15.12          & 46.01          & 50.04          & 0.88          & 13.07          & 46.79          & 50.74          & 0.61          & 10.14          & 35.48          & 26.23           \\
                                         & MemoryOS~\citeyearpar{kang-etal-2025-memory}                          & 1.14          & 15.46          & 45.73          & 47.87          & 0.97          & 13.86          & 47.55          & 46.28          & 0.69          & 10.33          & 35.87          & 24.58           \\
\cdashline{2-14}
                                         & MemoryBank~\citeyearpar{zhong2024memorybank}                        & 1.08          & 15.14          & 47.27          & 45.95          & 1.03          & 13.74          & \uline{48.39}  & 45.51          & 0.64          & 10.05          & 35.78          & 23.32           \\
                                         & LD-Agent~\citeyearpar{li-etal-2025-hello}                          & 1.37          & 15.78          & 46.42          & 50.16          & 1.02          & 14.05          & 47.76          & 48.63          & 0.72          & 10.47          & 35.96          & 25.94           \\
                                         & THEANINE~\citeyearpar{ong2024towards}                         & 1.27          & 14.84          & 45.69          & 54.23          & 0.94          & 13.55          & 47.42          & 53.64          & 0.79          & 10.23          & 35.77          & 28.97           \\
                                         
                                         & \multicolumn{13}{c}{{\cellcolor{Iron}}\textit{\textbf{Llama3-Instruct}}}                                                                                                                                                           \\
                                         & \textbf{\textsc{Icml}-1B}                           & 2.31          & 18.72          & 47.62          & 56.63          & 1.42          & 15.30          & 47.99          & 54.71          & 1.21          & 11.09          & 40.74          & 34.39           \\
                                         & \textbf{\textsc{Icml}-3B}                           & 2.37          & 18.78          & 47.65          & \textbf{61.66} & \textbf{1.49} & 15.38          & 48.02          & 57.39          & 1.20          & 11.21          & 40.80          & \uline{36.37}   \\
                                         & \textbf{\textsc{Icml}-8B}                           & 2.31          & 18.29          & 47.40          & 57.76          & \uline{1.46}  & 15.41          & 48.04          & \uline{57.53}  & \textbf{1.25} & 11.26          & 40.84          & 36.42           \\
                                         & \multicolumn{13}{c}{{\cellcolor{Iron}}\textit{\textbf{Gemma3-it}}}                                                                                                                                                                 \\
                                         & \textbf{\textsc{Icml}-1B}                           & 2.25          & 18.88          & \uline{47.76}  & 57.59          & 1.41          & 15.36          & 47.92          & 56.37          & 1.00          & 9.39           & 39.80          & 31.15           \\
                                         & \textbf{\textsc{Icml}-4B}                           & \textbf{2.44} & 18.88          & 47.70          & 58.39          & 1.36          & \uline{15.44}  & 48.01          & 57.02          & 1.16          & \uline{11.32}  & \textbf{40.87} & 35.19           \\
                                         & \textbf{\textsc{Icml}-12B}                          & 2.36          & 18.37          & 47.64          & \uline{58.77}  & 1.44          & 15.27          & 47.97          & 54.46          & 1.19          & 11.25          & 40.75          & 36.13           \\
                                         & \multicolumn{13}{c}{{\cellcolor{Iron}}\textit{\textbf{Qwen3}}}                                                                                                                                                                     \\
                                         & \textbf{\textsc{Icml}-1.7B}                         & 2.19          & \uline{18.92}  & 47.66          & 57.95          & 1.40          & 15.41          & 48.00          & 56.55          & \uline{1.23}  & 11.27          & 40.75          & \textbf{36.69}  \\
                                         & \textbf{\textsc{Icml}-4B}                           & 2.33          & 18.55          & 47.69          & 57.34          & 1.43          & 15.38          & 47.87          & 56.19          & 1.20          & 11.29          & 40.81          & 35.90           \\
                                         & \textbf{\textsc{Icml}-8B}                           & \uline{2.40}  & \textbf{18.93} & 47.74          & 57.60          & 1.40          & \textbf{15.45} & 47.96          & \textbf{57.58} & 1.17          & 11.25          & \uline{40.86}  & 35.85           \\ 
\hline
\multirow{19}{*}{\textbf{Gemini2.5}} & Long Context (1M)                  & 1.57          & 17.50          & 47.50          & 72.04          & 0.89          & 13.60          & 47.59          & 55.61          & 0.78          & 10.05          & 35.76          & 25.44           \\ 
\cdashline{2-14}
                                          & Mem0~\citeyearpar{chhikara2025mem0}                              & 1.09          & 15.88          & 44.73          & 52.94          & 0.93          & 12.92          & 45.88          & 49.46          & 0.66          & 9.64           & 35.67          & 26.93           \\
                                         & A-Mem~\citeyearpar{xu2025mem}                             & 1.18          & 14.97          & 45.38          & 51.86          & 0.75          & 12.42          & 46.85          & 49.71          & 0.76          & 10.28          & 36.12          & 25.94           \\
                                         & MemoryOS~\citeyearpar{kang-etal-2025-memory}                          & 1.26          & 15.75          & 45.97          & 51.93          & 0.91          & 12.78          & 45.76          & 48.94          & 0.71          & 9.53           & 35.58          & 26.47           \\
\cdashline{2-14}
                                         & MemoryBank~\citeyearpar{zhong2024memorybank}                        & 1.08          & 15.14          & 47.27          & 45.95          & 1.03          & 13.74          & \uline{48.39}  & 45.51          & 0.64          & 10.05          & 35.78          & 23.32           \\
                                         & LD-Agent~\citeyearpar{li-etal-2025-hello}                          & 1.43          & 16.17          & 45.78          & 60.42          & 0.96          & 12.54          & 45.47          & 49.93          & 0.74          & 9.83           & 35.49          & 27.96           \\
                                         & THEANINE~\citeyearpar{ong2024towards}                          & 1.64          & 17.02          & 45.23          & 75.42          & 1.07          & 14.27          & 46.01          & 55.64          & 0.91          & \textbf{11.45} & 36.98          & 30.29           \\
                                         
                                         & \multicolumn{13}{c}{{\cellcolor{Iron}}\textit{\textbf{Llama3-Instruct}}}                                                                                                                                                           \\
                                         & \textbf{\textsc{Icml}-1B}                           & 1.94          & 14.48          & 43.15          & 63.95          & 0.94          & 10.83          & 42.56          & 55.59          & 0.92          & 9.05           & 39.78          & 47.38           \\
                                         & \textbf{\textsc{Icml}-3B}                           & \textbf{2.47} & \textbf{18.97} & 47.64          & 78.45          & 1.12          & 13.78          & 46.13          & 64.23          & 0.90          & 9.58           & 39.79          & 43.69           \\
                                         & \textbf{\textsc{Icml}-8B}                           & \textbf{2.47} & 18.37          & 47.36          & 78.62          & \textbf{1.20} & 13.78          & 46.18          & 65.64          & 0.92          & 9.73           & \textbf{40.38}  & 50.42   \\
                                         & \multicolumn{13}{c}{{\cellcolor{Iron}}\textit{\textbf{Gemma3-it}}}                                                                                                                                                                 \\
                                         & \textbf{\textsc{Icml}-1B}                           & 1.74          & 14.47          & 43.56          & 66.39          & 1.11          & 13.65          & 46.00          & \uline{66.15}  & 0.65          & 8.12           & 38.99          & 43.13           \\
                                         & \textbf{\textsc{Icml}-4B}                           & 2.30          & 18.28          & 47.22          & 77.94          & \uline{1.13}  & \uline{13.85}  & 46.20          & \textbf{66.86} & 0.88          & 9.02           & 39.26          & 52.27  \\
                                         & \textbf{\textsc{Icml}-12B}                          & 2.39          & 17.94          & \textbf{47.85} & \uline{78.67}  & 1.07          & 13.77          & 46.12          & 65.89          & 0.77          & 8.37           & 39.40          & 42.10           \\
                                         & \multicolumn{13}{c}{{\cellcolor{Iron}}\textit{\textbf{Qwen3}}}                                                                                                                                                                     \\
                                         & \textbf{\textsc{Icml}-1.7B}                         & 1.98          & 17.25          & 46.15          & 75.48          & 1.07          & 13.56          & 42.55          & 65.50          & \uline{0.95}  & 9.96           & 39.99          & 54.38           \\
                                         & \textbf{\textsc{Icml}-4B}                           & \uline{2.42} & \uline{18.52}  & 47.70          & 77.60          & 1.05          & 13.33          & 46.29          & 65.68          & 0.89          & 10.22           & 39.89          & \uline{57.58}           \\
                                         & \textbf{\textsc{Icml}-8B}                           & 2.21          & 18.22          & \uline{47.82}  & \textbf{80.33} & \uline{1.13}  & \textbf{13.97} & \textbf{48.60} & 66.01          & \textbf{1.28} & \uline{10.64}  & \uline{40.20} & \textbf{59.81}           \\
\hline
\specialrule{1.5pt}{0pt}{0pt}
\end{tabular}}
\caption{Automatic evaluation (\%) of generation performance per episode. "\textbf{Bold Font}" means the highest results, while "\uline{Underlined Font}" means second-highest results.
*B-4 = BLEU-4, R-L = ROUGE-L, and Bert = BertScore. More results comparing memory-related methods and training reward curves are shown in Appendix~\ref{baselines} and~\ref{training_reward}.}
\label{auto}
\end{table*}

\section{Experiments}

\subsection{Experimental Settings}
Following~\citet{zhang2023mind} and~\citet{ong2024towards}, we evaluate our method on three long-term open-domain conversation datasets: \textbf{Multi-Session Chat} (MSC) \cite{xu2022beyond}, \textbf{Conversation Chronicles} (CC), \cite{jang2023conversation}, and \textbf{GapChat} (GC) \cite{zhang2023mind}. \textbf{These datasets comprise authentic human-to-human interactions, providing robust benchmarks to ensure generated responses align with real-world human expectations}. More details are shown in Appendix~\ref{datasetinfo}.

\paragraph{Models and Baselines.} For backbone, we evaluate on two closed-source long-context LLMs: 1) \noindent\textbf{GPT-4o (128K)}~\citep{hurst2024gpt}, the \texttt{gpt-4o-2024-11-20} version. 2) \textbf{Gemini2.5 (1M)}~\citep{comanici2025gemini}, the \texttt{gemini-2.5-pro-preview-03-25} version. For our method, we employ several state-of-the-art open-source LLMs: 1) \textbf{Llama-3.2 (1B/3B)} and \textbf{Llama-3.1 (8B)}, specifically the \texttt{-Instruct} versions. 2) \textbf{Gemma-3 (1B/4B/12B)}, using the \texttt{-it} versions. 3) \textbf{Qwen-3 (1.7B/4B/8B)}. We compare our \textbf{\textsc{Icml}} against various baselines. 1) \textbf{Long Context}: which use all the conversation histories. 2) \textbf{Management-centric memory agents}: \textbf{Mem0}~\cite{chhikara2025mem0}, \textbf{A-Mem}~\cite{xu2025mem}, and \textbf{MemoryOS}~\cite{ong2024towards}. 3) \textbf{Generation-centric dialogue agents}: \textbf{MemoryBank}~\citep{zhong2024memorybank}, 
\textbf{LD-Agent}~\citep{li-etal-2025-hello}, and \textbf{THEANINE}~\citep{ong2024towards}.
More details and baselines are shown in Appendix~\ref{baselines} and~\ref{Implementation Details}. Unless otherwise specified, we employ \textbf{Qwen3-8B} for training and \textbf{Gemini2.5} as the backbone in the following experiments and analyses.

\paragraph{Evaluation Metrics.}
We comprehensively evaluate our \textbf{\textsc{Icml}} on three types of metrics. 1) \textbf{Automatic Metrics.} Following \citet{ong2024towards}, we use BLEU-4 \citep{papineni2002bleu}, ROUGE-L \citep{lin2004rouge}, BertScore \citep{zhang2019bertscore}, and Mauve \citep{pillutla2021mauve} to automatically evaluate response generation. 2) \textbf{Personalized Metrics.} Following \citet{xu2022long} and \citet{jang2023conversation}, we introduce LLM-as-a-Judge~\cite{zheng2023judging} to evaluate response generation on five dimensions: \textit{Engagingness}, \textit{Humanness}, \textit{Coherence}, \textit{Consistency}, and \textit{Memorability}. 3) \textbf{Human Metrics.} Following \citet{xu2022long} and \citet{jang2023conversation}, we evaluate the winning performance of different methods on response generation and memory retrieval. More details of metrics are shown in Appendix~\ref{g-eval}.

\begin{figure}[!t]
\centering
\begin{subfigure}[b]{0.49\linewidth}
    \centering
    \includegraphics[width=\linewidth]{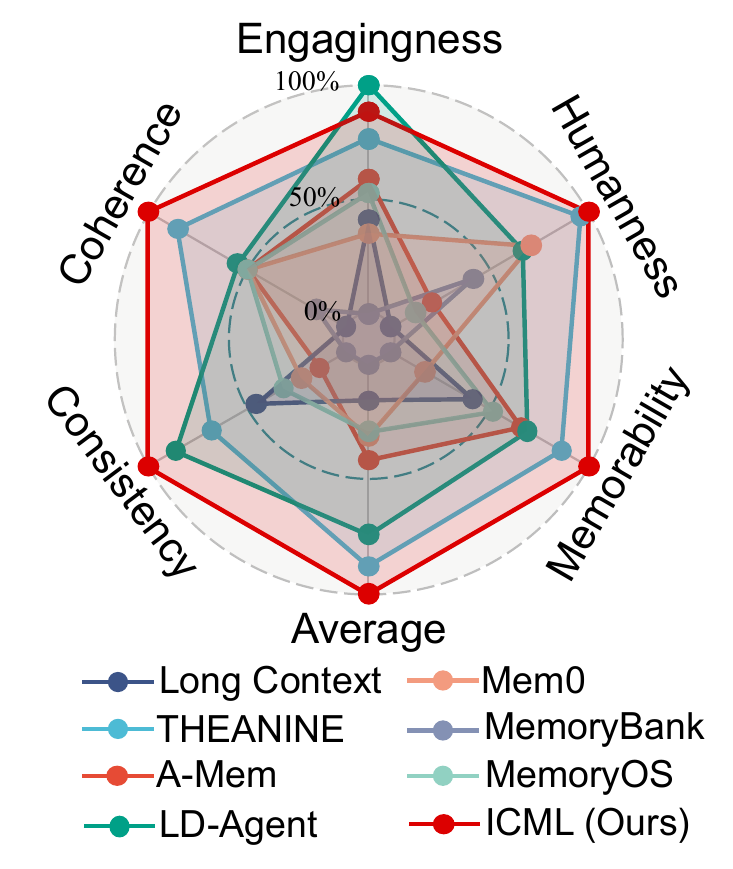}
    \caption{GPT-4o evaluation.}
    \label{GPT-4o evaluation}
\end{subfigure}
\hfill
\begin{subfigure}[b]{0.49\linewidth}
    \centering
    \includegraphics[width=\linewidth]{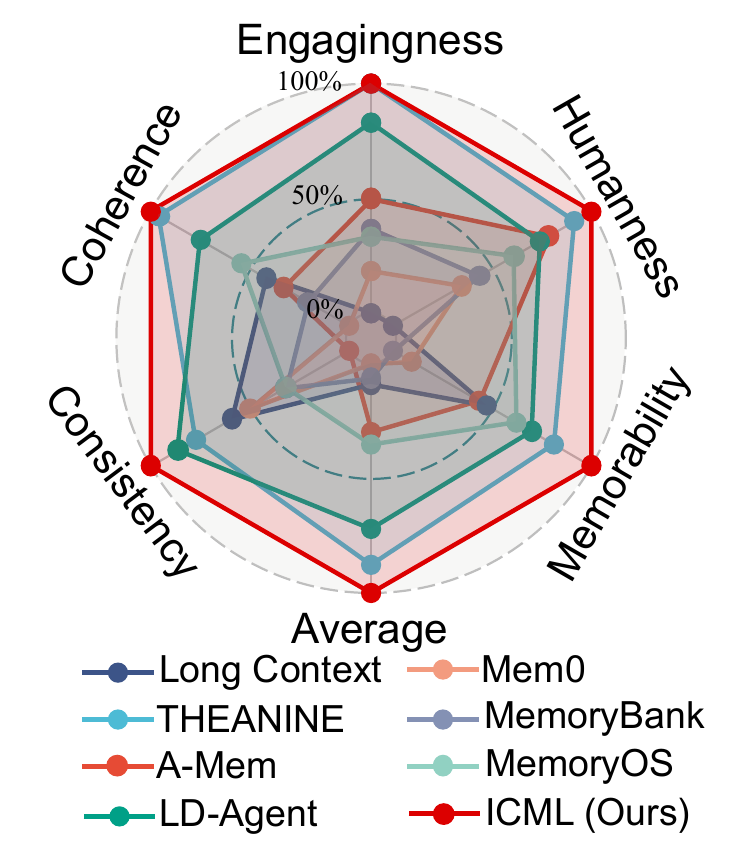}
    \caption{Gemini2.5 evaluation.}
    \label{Gemini2.5 evaluation}
\end{subfigure}
\caption{LLM cross-evaluation.}
\label{rader}
\end{figure}

\begin{figure}[!t]
\centering
\includegraphics[width=\linewidth]{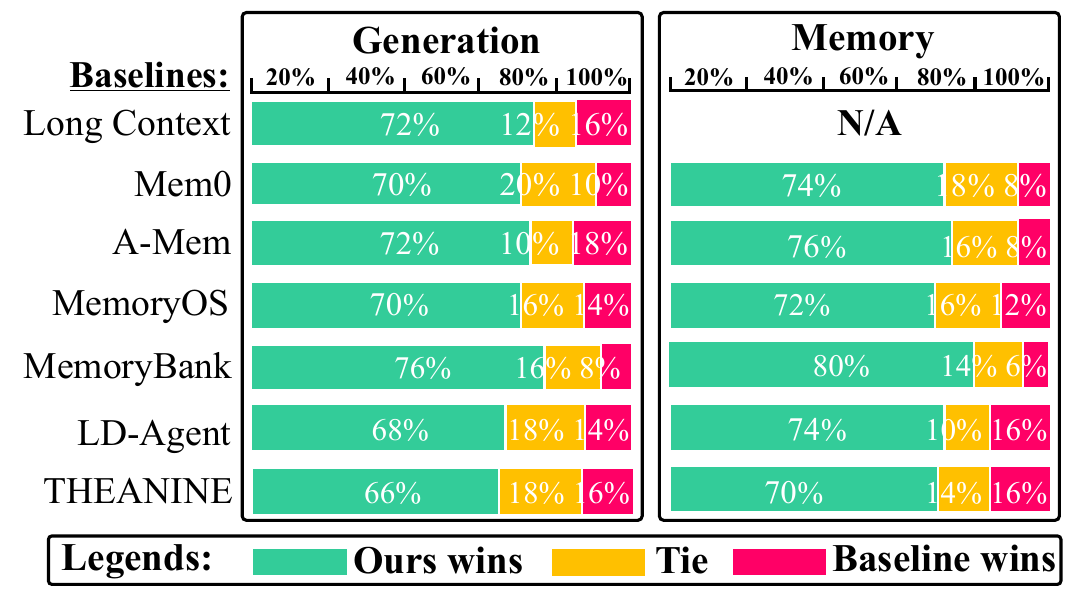}
\caption{Human evaluation on generation and memory.}
\label{human}
\end{figure}

\subsection{Main Results}
\paragraph{Evolving memory surpasses static heuristics.} 
Table~\ref{auto} shows that \textbf{\textsc{Icml}} achieves state-of-the-art results across all datasets, consistently outperforming management-centric memory agents, generation-centric dialogue agents, and long-context baselines. This proves that actively selecting high-value memories is far more effective than simply processing the entire, noise-filled history. Furthermore, our framework features a flexible plug-and-play design. It can be directly integrated with top-tier closed-source LLMs, equipping them with evolving long-term memory capabilities without requiring access to their internal weights.

\paragraph{Consistent alignment with LLMs and human expectations.}
The core insight from our subjective evaluations is that \textbf{\textsc{Icml}} achieves a unified consensus between automated model judgments and real human preferences. Unlike static baselines that often struggle to balance accurate recall with engaging conversation, our interactive learning paradigm effectively bridges this gap, delivering responses that are both contextually precise and naturally fluid. As illustrated in Figure~\ref{rader} and Figure~\ref{human}, this superiority is consistently verified: \textbf{\textsc{Icml}} not only demonstrates comprehensive improvements across all dimensions in LLM cross-evaluation but also secures a dominant preference in human evaluation. This confirms that evolving memory policies through interaction leads to a generation style that is significantly more attuned to user expectations than traditional methods.

\begin{table}[!t]
\renewcommand{\arraystretch}{1.2}
\small
\centering
\setlength{\tabcolsep}{1mm}{
\begin{tabular}{cccccc} 
\specialrule{1.5pt}{0pt}{0pt}
\hline
\textbf{Datasets}             & \textbf{Methods} & \textbf{B-4}  & \textbf{R-L}   & \textbf{Bert}  & \textbf{Mauve}  \\ 
\hline
\multirow{6}{*}{\textbf{CC}}  & \cellcolor{Iron}\textbf{\textsc{Icml}} (Ours)    & \cellcolor{Iron}\textbf{2.21} & \cellcolor{Iron}\textbf{18.22} & \cellcolor{Iron}\textbf{47.82} & \cellcolor{Iron}\textbf{80.33}  \\ 
\cdashline{2-6}
                              & w/o Synthetic Data   & 2.08          & 18.15          & 46.87          & 79.57           \\
\cdashline{2-6}
                              & w/o Planner Agent & 2.01          & 18.01          & 47.04          & 78.77           \\
                              & w/o Trigger Agent & 2.10          & 18.18          & 46.83          & 79.91           \\
                              & w/o Truth Reward & 1.95          & 17.96          & 47.15          & 76.19           \\
\cdashline{2-6}
                              & w/o Evolution & 2.16          & 18.14          & 47.31          & 79.83           \\
\hline
\multirow{6}{*}{\textbf{MSC}} & \cellcolor{Iron}\textbf{\textsc{Icml}} (Ours)    & \cellcolor{Iron}\textbf{1.13} & \cellcolor{Iron}\textbf{13.97} & \cellcolor{Iron}\textbf{48.60} & \cellcolor{Iron}\textbf{66.01}  \\ 
\cdashline{2-6}
                              & w/o Synthetic Data   & 1.01          & 13.61          & 45.85          & 62.56           \\
\cdashline{2-6}
                              & w/o Planner Agent & 1.05          & 13.46          & 45.79          & 63.54           \\
                              & w/o Trigger Agent & 1.08          & 13.31          & 45.44          & 62.62           \\
                              & w/o Truth Reward & 1.10          & 13.62          & 46.00          & 65.99           \\
\cdashline{2-6}
                              & w/o Evolution & 1.03          & 13.86          & 46.33          & 61.81           \\
\hline
\multirow{6}{*}{\textbf{GC}}  & \cellcolor{Iron}\textbf{\textsc{Icml}} (Ours)    & \cellcolor{Iron}\textbf{1.28} & \cellcolor{Iron}\textbf{10.64} & \cellcolor{Iron}\textbf{40.20} & \cellcolor{Iron}\textbf{59.81}  \\ 
\cdashline{2-6}
                              & w/o Synthetic Data   & \textbf{1.28}          & 10.47          & 40.18          & 59.07           \\
\cdashline{2-6}
                              & w/o Planner Agent & 1.26          & 10.07          & 40.09          & 58.07           \\
                              & w/o Trigger Agent & 1.22          & 10.25          & 40.02          & 58.35           \\
                              & w/o Truth Reward & 0.97          & 10.35          & 40.11          & 58.78           \\
\cdashline{2-6}
                              & w/o Evolution & 1.16          & 9.96          & 40.26          & 59.08           \\
\hline
\specialrule{1.5pt}{0pt}{0pt}
\end{tabular}}
\caption{Ablation study of our method. We further investigate different reward models in Appendix~\ref{Robustness}.}
\label{ablation}
\end{table}

\paragraph{Holistic integrity sustains the evolutionary cycle.}

Table~\ref{ablation} confirms that every component is essential for optimal performance. Removing the \textbf{Cross-Session Truth Reward} causes the sharpest drop, proving that long-term feedback is vital for judging memory utility. Similarly, the decline without \textbf{Evolution} shows that static training is insufficient, and the agent must adapt continuously during testing. \textbf{Synthetic Data} is also critical, as it solves the cold-start problem by providing initial expert examples. Finally, removing either the \texttt{Planner} or \texttt{Trigger} breaks the collaborative workflow, confirming that both agents must cooperate for effective memory management.

\subsection{Analysis of Collaborative Agents}

\begin{figure}[!t]
\centering
\begin{subfigure}[b]{0.494\linewidth}
    \centering
    \includegraphics[width=\linewidth]{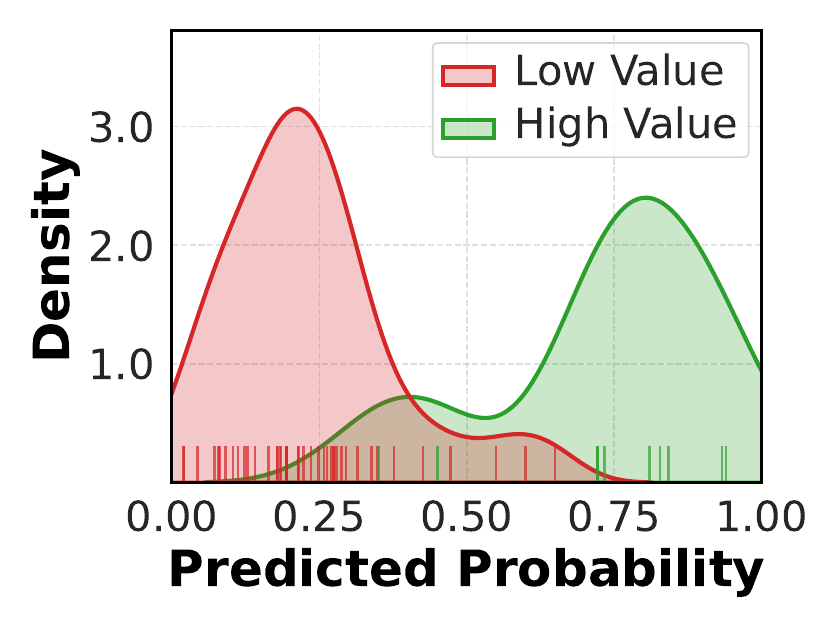}
    \caption{Value discrimination.}
    \label{fig:actor1}
\end{subfigure}
\hfill
\begin{subfigure}[b]{0.494\linewidth}
    \centering
    \includegraphics[width=\linewidth]{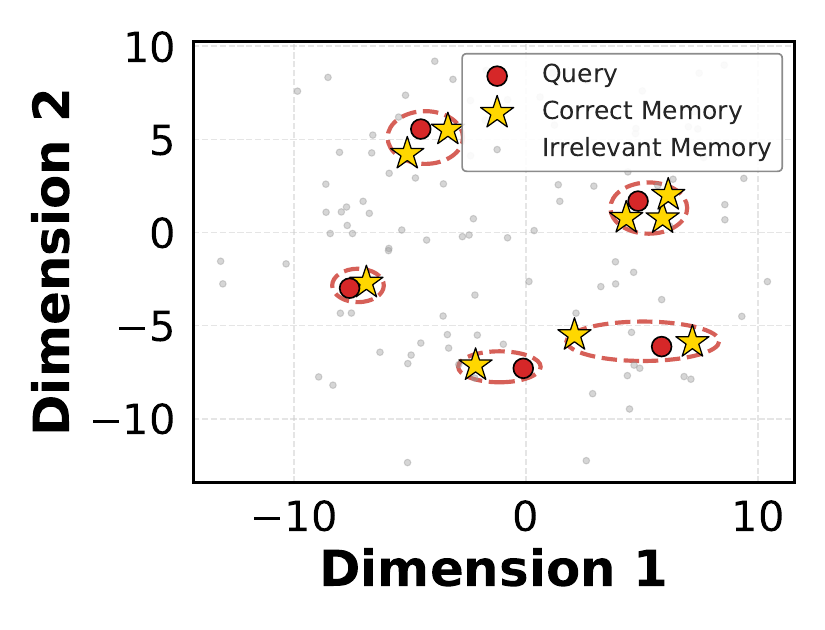}
    \caption{Retrieval space.}
    \label{fig:actor2}
\end{subfigure}
\caption{Visualization of learned \texttt{Planner} policy (Left) and \texttt{Trigger} policy (Right).}
\label{fig:actors_comparison}
\end{figure}

\paragraph{Emergence of distinct decision boundaries.} To intuitively understand the learned policies, we visualize the decision landscapes of both agents from an episode in Figure~\ref{fig:actors_comparison}. As shown in Figure~\ref{fig:actors_comparison} (a), the \texttt{Planner} develops a sharp discrimination ability after evolution, where it assigns distinctively high probabilities to valuable information while effectively suppressing low-value noise. Complementing this, the t-SNE visualization of the \texttt{Trigger} in Figure~\ref{fig:actors_comparison} (b) reveals that user queries and ground-truth memories form tight semantic clusters separated from irrelevant noise. This spatial alignment confirms that the agent has successfully learned to map current user needs to precise historical contexts, ensuring accurate retrieval even in complex scenarios.

\begin{figure}[!t]
\centering
\includegraphics[width=\linewidth]{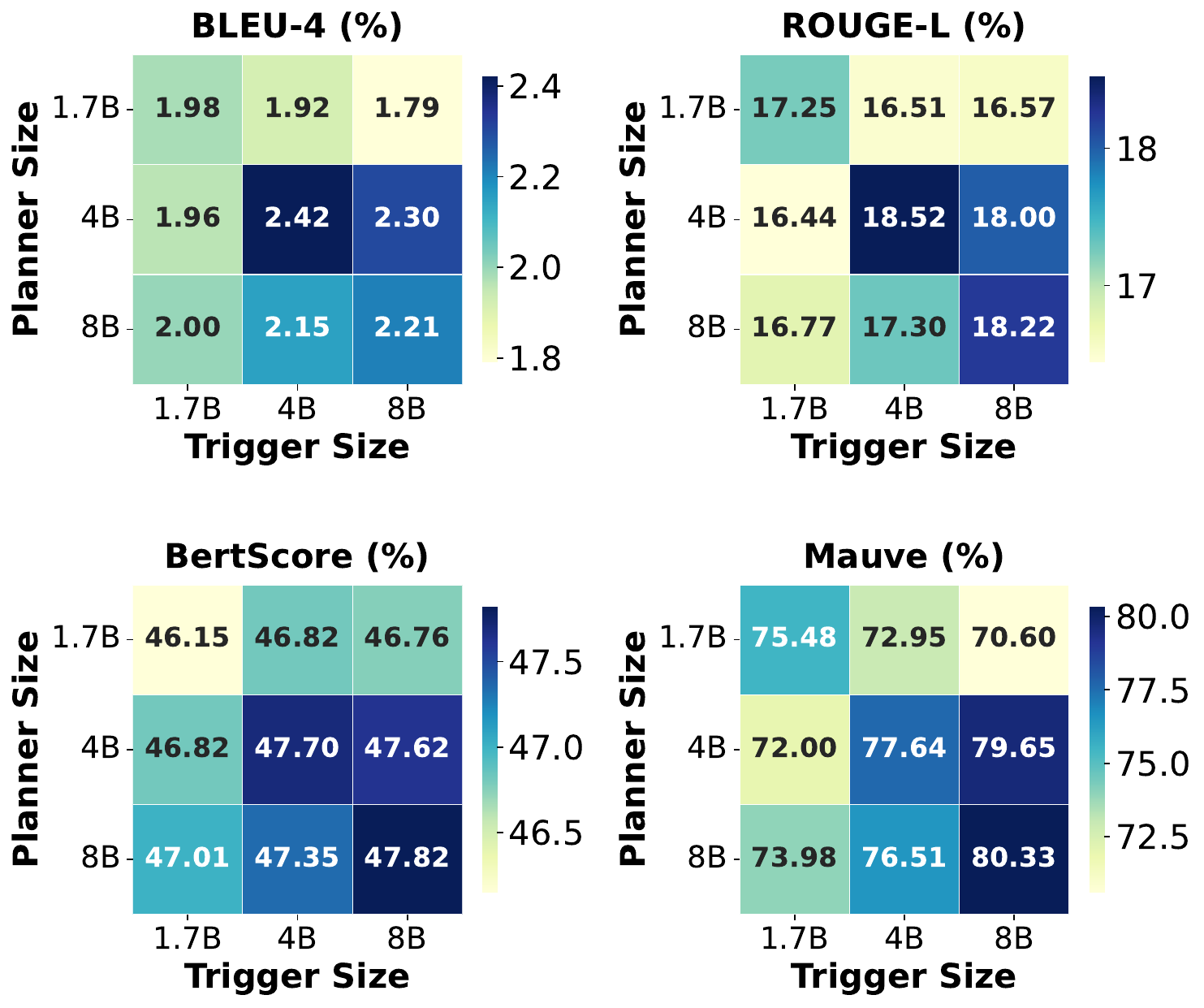}
\caption{Performance scaling of \texttt{Planner} and \texttt{Trigger} on CC dataset. More results are shown in Appendix~\ref{scaling_msc_gc}.}
\label{scaling}
\end{figure}

\begin{figure}[!t]
\centering
\includegraphics[width=\linewidth]{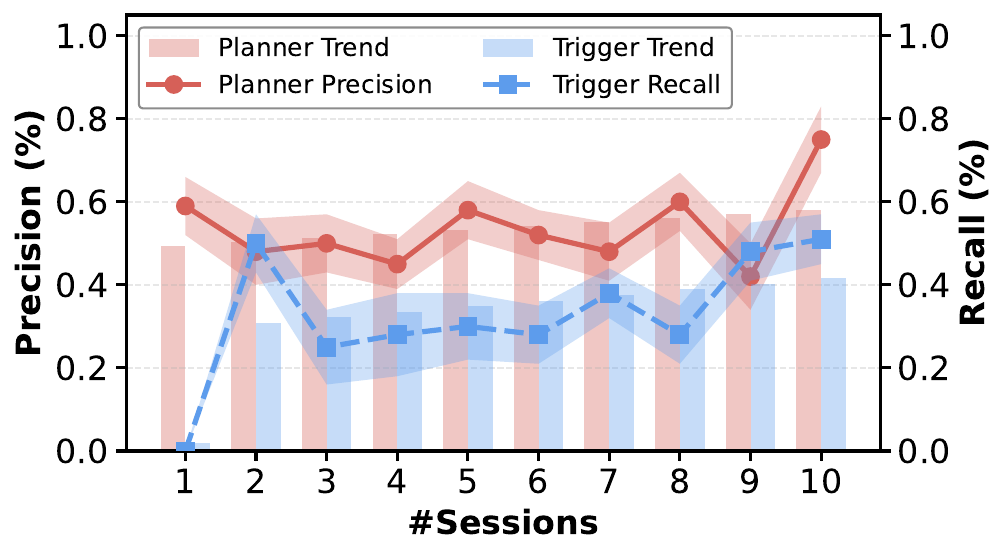}
\caption{Co-evolution of \texttt{Planner} and \texttt{Trigger} performance over online sessions.}
\label{coevolution}
\end{figure}

\begin{table}[!t]
\renewcommand{\arraystretch}{1.2}
\small
\centering
\setlength{\tabcolsep}{1.2mm}{
\begin{tabular}{cccccc} 
\specialrule{1.5pt}{0pt}{0pt}
\hline
\textbf{Datasets}             & \textbf{Synthetic Size} & \textbf{B-4}  & \textbf{R-L}   & \textbf{Bert}  & \textbf{Mauve}  \\ 
\hline
\multirow{5}{*}{\textbf{CC}}  & \cellcolor{Iron}None                    & \cellcolor{Iron}2.08          & \cellcolor{Iron}18.15          & \cellcolor{Iron}46.87          & \cellcolor{Iron}79.57           \\ 
\cdashline{2-6}
                              & 0.25K                   & 2.21          & \uline{18.22}          & \textbf{47.82} & \textbf{80.33}  \\
                              & 0.5K                    & \textbf{2.62} & \textbf{18.59} & \uline{47.28}          & \uline{79.60}           \\
                              & 0.75K                   & \uline{2.32}          & 17.92          & 46.83          & 77.87           \\
                              & 1K                      & 2.02          & 17.24          & 46.57          & 76.33           \\ 
\hline
\multirow{5}{*}{\textbf{MSC}} & \cellcolor{Iron}None                    & \cellcolor{Iron}1.01          & \cellcolor{Iron}13.61          & \cellcolor{Iron}45.85          & \cellcolor{Iron}62.56           \\ 
\cdashline{2-6}
                              & 0.25K                   & \uline{1.13}          & \textbf{13.97} & \textbf{48.60} & \uline{66.01}           \\
                              & 0.5K                    & \textbf{1.18} & \uline{13.67}          & \uline{46.06}          & \textbf{66.45}  \\
                              & 0.75K                   & 1.10          & 13.42          & 45.86          & 65.25           \\
                              & 1K                      & 1.05          & 13.56          & 45.81          & 65.89           \\ 
\hline
\multirow{5}{*}{\textbf{GC}}  & \cellcolor{Iron}None                    & \cellcolor{Iron}\textbf{1.28} & \cellcolor{Iron}\uline{10.47}          & \cellcolor{Iron}40.18          & \cellcolor{Iron}\uline{59.07}           \\ 
\cdashline{2-6}
                              & 0.25K                   & \textbf{1.28} & \textbf{10.64} & 40.20          & \textbf{59.81}  \\
                              & 0.5K                    & 0.90          & 10.04          & \uline{40.22}          & 49.21           \\
                              & 0.75K                   & 0.93          & 10.23          & \textbf{40.37} & 50.30           \\
                              & 1K                      & \uline{0.95}          & 10.42          & 40.31          & 49.39           \\
\hline
\specialrule{1.5pt}{0pt}{0pt}
\end{tabular}}
\caption{Performance scaling with synthetic data size.}
\label{datasize}
\end{table}

\begin{figure*}[!t]
\centering
\includegraphics[width=\textwidth]{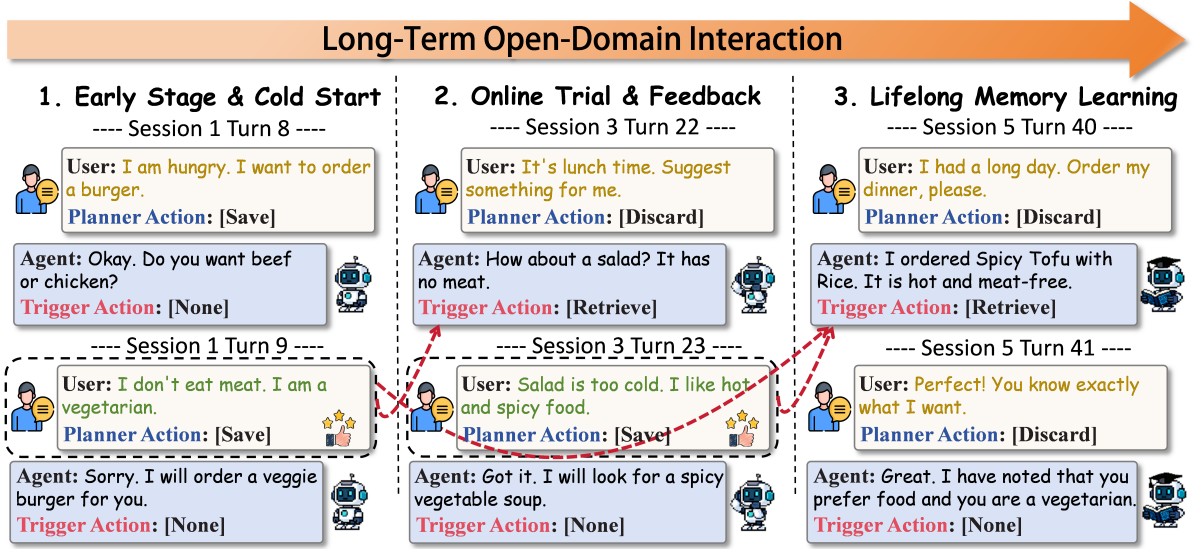}
\caption{Case study of the interactive memory learning process. The \textcolor[RGB]{200,29,49}{red dashed arrow} shows memory retrieval. The ultimate evolutionary goal is further illustrated in Figure~\ref{intro_extra} (Appendix~\ref{expectation}).}
\label{case}
\end{figure*}

\begin{figure}[!t]
\centering
\includegraphics[width=\linewidth]{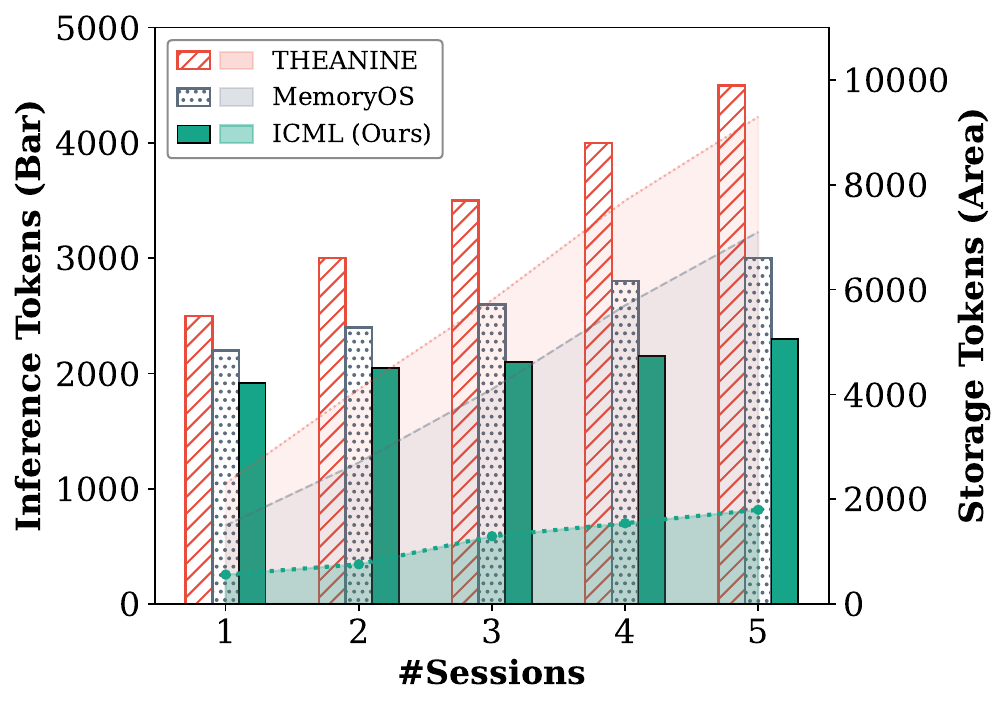}
\caption{Inference tokens and storage tokens.}
\label{tokens}
\end{figure}

\paragraph{Balanced scaling facilitates efficient collaboration.}
We examine the impact of model scaling in Figure~\ref{scaling}. In most cases, the results exhibit a diagonal pattern where performance peaks with matched model sizes, suggesting that aligned capabilities facilitate the collaborative loop. However, larger models also contribute positive gains due to their enhanced raw capacity. Notably, smaller but paired models frequently yield competitive results against mismatched configurations, indicating that architectural balance is often a cost-effective strategy for maximizing synergy.

\paragraph{Continuous improvement via co-evolution.}
To evaluate lifelong adaptation capabilities, we extend the interaction to 10 sessions and label ground truths as shown in Figure~\ref{coevolution}. The results demonstrate a consistent upward trend for both \texttt{Planner} precision and \texttt{Trigger} recall as the dialogue progresses. This confirms the effectiveness of our self-evolutionary mechanism: the agents actively refine their collaborative strategies through continuous environmental feedback, progressively enhancing their coordination to sustain high-quality generation over interactions.

\paragraph{Moderate warm-up enables test-time adaptation.}
We explore the scaling effects of synthetic data in Table~\ref{datasize}. The results indicate that a modest range of 0.25K to 0.5K episodes yields the optimal performance gain. This phenomenon stems from the constantly changing nature of user expectations: insufficient data fails to overcome the cold-start problem, hindering rapid adaptation; conversely, excessive static supervision risks overfitting to fixed patterns, reducing the agent's flexibility to align with shifting real-time preferences. Therefore, a moderate warm-up strikes the best balance, initializing the policy just enough to unlock \textbf{\textsc{Icml}}'s capability for autonomous test-time evolution.

\subsection{Case Study}
Figure~\ref{case} illustrates how \textbf{\textsc{Icml}} evolves through real-time interaction. Initially capturing the "vegetarian" constraint, the agent later encounters a conflict when the user rejects a cold salad. Instead of failing, the \texttt{Planner} adaptively updates its memory to include the specific "hot and spicy" preference derived from this feedback. Consequently, the \texttt{Trigger} successfully synthesizes both the long-term restriction and the newly learned preference to recommend "Spicy Tofu", perfectly aligning with the user's expectations.

\begin{figure}[!t]
\centering
\begin{subfigure}[b]{0.49\linewidth}
    \centering
    \includegraphics[width=\linewidth]{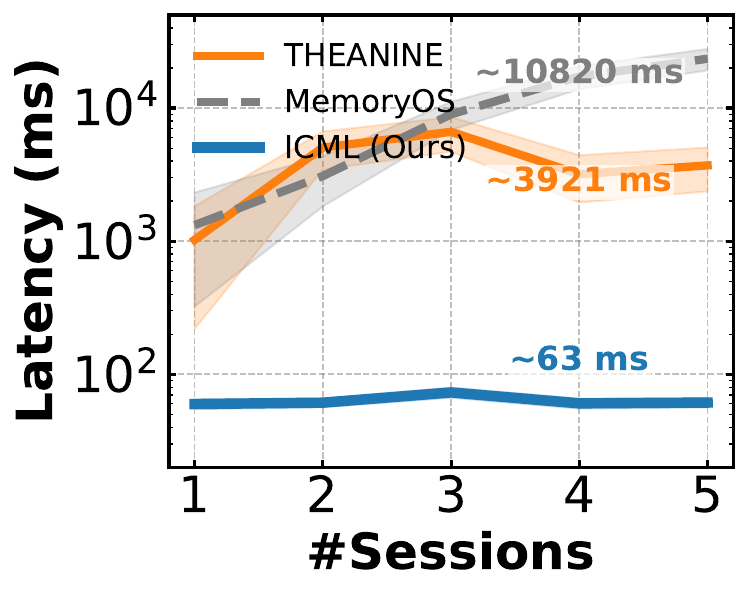}
    \caption{Construction latency.}
    \label{latency_construction}
\end{subfigure}
\hfill
\begin{subfigure}[b]{0.49\linewidth}
    \centering
    \includegraphics[width=\linewidth]{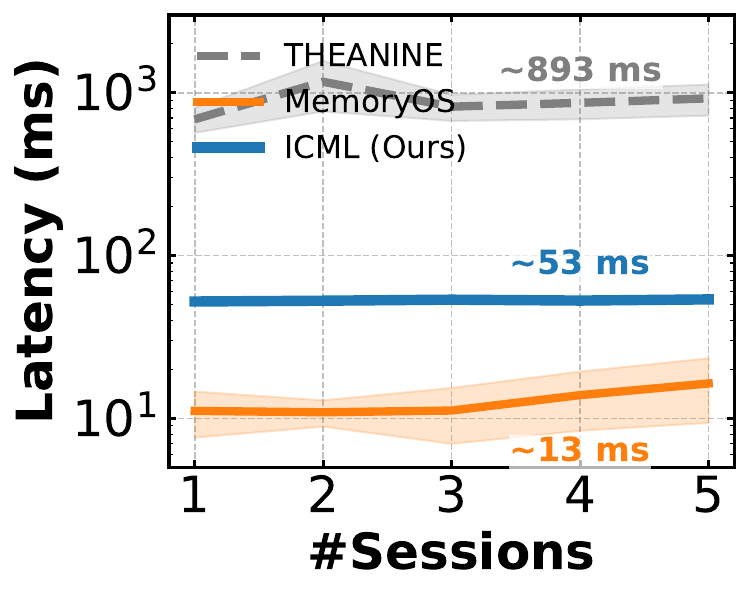}
    \caption{Retrieval latency.}
    \label{latency_retrieval}
\end{subfigure}
\caption{Computational time cost. More results of total processing time are shown in Appendix~\ref{session_efficiency}.}
\label{latency}
\end{figure}

\subsection{Analysis of Token and Latency Efficiency}
\paragraph{Token efficiency.}
We analyze the token consumption in Figure~\ref{tokens}. Unlike baselines where costs escalate linearly with session depth, \textbf{\textsc{Icml}} maintains remarkably stable inference usage (bars) and minimal storage growth (area). This proves that the \texttt{Planner}'s rigorous noise filtering effectively prevents context bloating, ensuring that long-term interaction remains computationally feasible without sacrificing performance.

\paragraph{Latency efficiency.}

As shown in Figure~\ref{latency}, \textbf{\textsc{Icml}} achieves fast memory construction, performing much better than other methods that rely on complex processing. This improvement removes the main delay in the system. This ensures our system is ready for real-time use where quick response generation is needed.

\section{Conclusions}

In this paper, we present \textbf{\textsc{Icml}}, a collaborative framework where a \texttt{Planner} and \texttt{Trigger} co-evolve to optimize long-term memory. By leveraging delayed feedback, our approach effectively aligns memory operations with actual conversational utility, ensuring the agent retains only truly valuable information. Extensive experiments demonstrate that \textbf{\textsc{Icml}} significantly outperforms strong baselines in generation quality while maintaining millisecond-level latency and stable token consumption. Furthermore, our analysis confirms that the system achieves continuous self-evolution through online interaction, offering a robust and efficient solution for lifelong personalized assistants.
\section*{Limitations}

Our work is dedicated to constructing personal conversational assistants capable of self-evolution through deep, long-term open-domain interaction. Consequently, our evaluation prioritizes open-domain engagement and personalized alignment rather than rigid reasoning or strict fact-retrieval tasks, such as complex mathematics, coding, or standard question answering benchmarks, which lie beyond the scope of this companionship-centric goal. Furthermore, while we validate our approach within the dialogue domain, we propose a novel paradigm for interactive memory learning. We believe this framework offers valuable insights into dynamic information retention, with the potential to inspire future adaptations across broader domains involving complex temporal dependencies.
\section*{Acknowledgements}
This work was supported by the National Natural Science Foundation of China 62576120 and the Major Key Project of PCL2025A11 and PCL2024A08. Thanks for the support provided by OpenI Community (https://openi.pcl.ac.cn).

\bibliography{anthology,custom}
\bibliographystyle{acl_natbib}
\appendix

\begin{table*}[!t]
\centering
\small
\renewcommand{\arraystretch}{1.2}
\begin{tabular}{cccccc}
\specialrule{1.5pt}{0pt}{0pt}
\textbf{Datasets} & \textbf{\# of Sessions} & \textbf{\# of Episodes} & \textbf{\# of Turns} & \textbf{Avg. Turns per Session} & \textbf{Avg. Turns per Episode} \\
\hline
\textbf{CC}       & 1M             & 200K           & 11.7M       & 11.70                  & 58.50                  \\
\textbf{MSC}      & 16K            & 5K             & 214K        & 13.38                  & 42.80                  \\
\textbf{GC}       & 2.65K          & 0.65K          & 28.13K      & 10.62                  & 43.28                  \\
\specialrule{1.5pt}{0pt}{0pt}
\end{tabular}
\caption{The statistics of three long-term open domain datasets.}
\label{dataset}
\end{table*}

\section{Dataset Information}
\label{datasetinfo}
We evaluate our method on three long-term multi-session conversation datasets: \textbf{Conversation Chronicles} (CC)~\cite{jang2023conversation}, \textbf{Multi-Session Chat} (MSC)~\cite{xu2022beyond}, and \textbf{GapChat} (GC)~\cite{zhang2023mind}:
\begin{itemize}
\item[$\bullet$] \textbf{CC}: It features a 1M multi-session dialogue dataset that emphasizes temporal dynamics and complex speaker relationships in long-term interactions. It captures the natural flow and logical development found in real human conversations, maintaining coherent and consistent interactions across many sessions. This dataset helps agents learn how to communicate naturally and personally, just like humans do in diverse social contexts.
\item[$\bullet$] \textbf{MSC}: It is a large-scale, long-term open-domain dialogue dataset built from authentic human-to-human interactions across multiple sessions. In this dataset, speakers learn about each other's interests over time and discuss things they have learned in past conversations. It mimics the way real humans build relationships through long-term interaction, making it a key benchmark for testing an agent's long-term memory.
\item[$\bullet$] \textbf{GC}: It is a challenging multi-session dialogue dataset that incorporates realistic time intervals between conversations, ranging from minutes to years. To create realistic long-term dialogues, it simulates progress in the speakers' lives based on real-world human rhythms. This dataset requires agents to perceive the passage of time like humans and accurately adapt to changes in a user’s life across different session gaps.
\end{itemize}

These datasets are human-verified and built through a meticulous crowdsourcing pipeline, purpose-built for the simulation and evaluation of long-term, context-dependent conversations. Following~\citet{ong2024towards}, we randomly select 50 episodes from the test set of each dataset, a total of 250 sessions for the experiments in this paper. The statistics of each data set are shown in Table \ref{dataset}.%

\section{Compared Baselines}
\label{baselines}
To evaluate the effectiveness of our approach, we compare it against two primary categories of baselines: management-centric memory agents and generation-centric dialogue agents.

\subsection{Management-Centric Memory Agents} This category focuses on the autonomous organization and structural maintenance of the memory database:
\begin{itemize}
\item[$\bullet$] \textbf{Mem0}~\cite{chhikara2025mem0}: Memo uses a scalable architecture with two phases: extraction and update. In the extraction phase, the system picks out key facts from conversation pairs. In the update phase, it manages memory using a tool call to decide whether to add, update, delete, or ignore new information.
\item[$\bullet$] \textbf{A-Mem}~\cite{xu2025mem}: A-Mem is an agentic memory system inspired by the Zettelkasten method. It turns conversation steps into atomic notes that include keywords, tags, and context descriptions. The system finds relevant past notes by comparing their embedding vectors.
\item[$\bullet$] \textbf{MemoryOS}~\cite{kang-etal-2025-memory}: MemoryOS is inspired by operating systems and uses three levels of storage: short-term, mid-term, and long-term personal memory. It organizes data using a segment-page strategy, where dialogues about the same topic are grouped into segments and divided into pages.
\end{itemize}

\subsection{Generation-Centric Dialogue Agents} Beyond memory management, we also compare our method with state-of-the-art agents that prioritize long-term consistency and personalized response generation:
\begin{itemize}
\item[$\bullet$] \textbf{MemoryBank}~\cite{zhong2024memorybank}: MemoryBank provides a long-term memory system for LLMs inspired by human memory. It saves conversation logs and summarizes them into hierarchical event summaries, allowing the agent to better adapt to the user's personality.
\item[$\bullet$] \textbf{LD-Agent}~\cite{li-etal-2025-hello}: This paper introduces a framework called LD-Agent for personalized, long-term dialogue. The system uses a modular design, breaking the task into three separate parts: event perception, persona extraction, and response generation.
\item[$\bullet$] \textbf{THEANINE}~\cite{ong2024towards}: THEANINE helps dialogue agents manage memory without deleting old information. While most systems throw away old data, THEANINE keeps everything because it believes even outdated information provides important context, such as changes in user behavior.
\end{itemize}

\subsection{Memory-Related Methods}
Furthermore, we compare our work with a range of established memory-related methods. These approaches typically focus on enhancing memory utilization through recursive summarization or instruction-based techniques to support long-term dialogues:
\begin{itemize}
\item[$\bullet$] \textbf{MemoChat}~\cite{lu2023memochat}: MemoChat uses instruction tuning to help models maintain consistency in long conversations via self-composed "memos". It follows a cycle of "memorization-retrieval-response" to ensure the agent effectively uses historical information.
\item[$\bullet$] \textbf{Rsum}~\cite{wang2025recursively}: Rsum proposes a recursive summarization mechanism. It guides the model to first memorize small dialogue segments and then recursively generate new memory by combining old memory with the subsequent context to maintain consistency over time.
\item[$\bullet$] \textbf{COMEDY}~\cite{chen2025compress}: COMEDY moves away from traditional retrieval modules and uses a single model for memory generation, compression, and response. It integrates dialogue summaries, user-bot dynamics, and past events into a concise "compressive memory" format.
\end{itemize}

\section{Implementation Details}\label{Implementation Details}
To ensure the reproducibility of our \textbf{\textsc{Icml}} framework, we summarize the key hyperparameters used in both the supervised warm-up and the online reinforcement learning stages in Table~\ref{tab:hyperparameters}.

\begin{table}[!t]
\small
\centering
\renewcommand{\arraystretch}{1.2}
\setlength{\tabcolsep}{1.2mm}{
\begin{tabular}{lll}
\specialrule{1.5pt}{0pt}{0pt}
\toprule
\textbf{Category} & \textbf{Hyperparameter} & \textbf{Value} \\ 
\midrule
\multirow{3}{*}{Data Synthesis} 
    & Storylines per Episode & 3 \\
    & Prequels per Storyline & 4 \\
    & Data Size & 0.25K \\
\midrule
\multirow{3}{*}{Cold Start} 
    & Learning Rate & $2 \times 10^{-5}$ \\
    & Training Epochs & 3 \\
    & Batch Size & 1/2/4/8/16 \\
\midrule
\multirow{7}{*}{RL} 
    & Actor Learning Rate & $1 \times 10^{-6}$ \\
    & Critic Learning Rate & $1 \times 10^{-5}$ \\
    & Discount Factor ($\gamma$) & 0.99 \\
    & GAE Lambda ($\lambda$) & 0.95 \\
    & PPO Clip Epsilon ($\epsilon$) & 0.2 \\
    & Reward Normalization& [0, 1] \\ 
    & Batch Size & 4 \\
\midrule
\multirow{4}{*}{LoRA} 
    & rank & 8 \\
    & lora\_alpha & 16 \\
    & lora\_dropout & 0.1 \\
    & bias & none \\
\bottomrule
\specialrule{1.5pt}{0pt}{0pt}
\end{tabular}}
\caption{Key hyperparameters for \textbf{\textsc{Icml}} training.}
\label{tab:hyperparameters}
\end{table}

\begin{algorithm}[!t]
\small
\caption{Interactive Memory Learning}\label{alg:icml}
\SetAlgoLined
\KwIn{Expert data $\mathcal{D}_{expert}$, discount $\gamma$, learning rates $\eta$}
\KwOut{Optimized policy $\pi_\theta$}

Pre-train $\pi_\theta$ on $\mathcal{D}_{expert}$\;
$\mathcal{B}_{pending} \leftarrow \emptyset$\;

\For{each episode $E$}{
    \For{each session $S \in E$}{
        \For{each turn $k \in S$}{
            $o_k = (u_k, H_k, \mathcal{M}_k)$\;
            $a_k^t \sim \pi_\theta^t(a_k^t|o_k)$ \tcp*[r]{Trigger Decision}
            \If{$a_k^t > 0$}{
                Generate $r^*$ using $m_{a_k^t}$ and obtain $r^{qual}_{k}$\;
                \If{$m_{a_k^t} \in \mathcal{B}_{pending}$}{
                    $r^{truth} \leftarrow r^{qual}_{k}$ \tcp*[r]{Reward Propagation}
                    $\mathcal{B}_{pending} \leftarrow \mathcal{B}_{pending} \setminus \{m_{a_k^t}\}$\;
                }
            }
            $a_k^p \sim \pi_\theta^p(a_k^p|o_k)$ \tcp*[r]{Planner Decision}
            \If{$a_k^p = 1$}{
                $\mathcal{M} \leftarrow \mathcal{M} \cup \{m_{new}\}$; Obtain $r^{proxy}_{k}$\;
                $\mathcal{B}_{pending} \leftarrow \mathcal{B}_{pending} \cup \{(o_k, a_k^p)\}$\;
            }
            Collect trajectory $\tau_k = \{o_k, a_k, r_k\}$\;
        }
        $\pi_\theta \leftarrow \text{PPO}(\pi_\theta, \tau, \eta)$ \tcp*[r]{Online Optimization}
    }
}
\end{algorithm}

\subsection{Supervised Warm-up Stage}
To mitigate the cold-start problem inherent in interactive learning, we first conduct supervised pre-training on expert trajectories generated via Retrospective Session Synthesis. In this phase, we employ a learning rate of $2 \times 10^{-5}$ for 3 epochs. The batch size is set to 1/2/4/8/16 (according to VRAM), and we employ LoRA~\cite{hu2022lora} to train our method.

\subsection{Interactive Reinforcement Learning Stage}
In the online self-evolution phase, we jointly optimize the \texttt{Planner} and \texttt{Trigger} modules using the PPO algorithm. The learning rate for the Actor is set to a relatively small value of $1 \times 10^{-6}$ to preserve policy stability, while the Critic uses $1 \times 10^{-5}$ to accelerate the convergence of the value function. We set the discount factor $\gamma = 0.99$, and utilize GAE~\cite{schulman2015high} ($\lambda=0.95$) alongside a clipping coefficient $\epsilon=0.2$ to balance bias and variance. All reward signals, derived from an LLM-as-a-Judge, are normalized within the range of [0, 1].

\section{Metrics}
\label{g-eval}
\subsection{Personalized Metrics}
With the development of open-domain conversation based on LLM, traditional overlap metrics such as BLEU \citep{papineni2002bleu}, ROUGE \citep{lin2004rouge}, etc. face great challenges. The reason is that a wide range of response generation can be considered as appropriate responses \citep{liu2016not}. To this end, we refer to LLM-as-a-Judge \citep{zheng2023judging} and use LLMs to evaluate episodes. In our paper, we follow the metrics set in \citet{xu2022long} and \citet{jang2023conversation}:

\begin{itemize}
\item[$\bullet$] \textbf{Engagingness}: The assistant can have rich interactions with users that go beyond simple conversations. For example, the assistant can generate interesting and immersive responses based on the current context.
\item[$\bullet$] \textbf{Humanness}: Measures the extent to which the assistant exhibits anthropomorphic traits. This includes the capacity for empathetic reasoning and the simulation of humanlike cognitive patterns during communication.
\item[$\bullet$] \textbf{Coherence}: Measures the logical and thematic continuity across both immediate turns and distant historical sessions. The assistant must synthesize information from different points in time to ensure the conversation flows naturally without losing the "thread".
\item[$\bullet$] \textbf{Consistency}: Focuses on the internal stability of the agent’s persona and its knowledge of the user. The assistant must avoid self-contradiction and maintain a persistent identity across interactions spanning days or weeks.
\item[$\bullet$] \textbf{Memorability}: Reflects the efficiency of the memory system in identifying and retrieving salient facts from past experiences. It evaluates whether the agent can correctly reference specific details, preferences, or events mentioned in earlier sessions to build a sense of shared history.
\end{itemize}
Each metric is scored on a scale of 1-5, with 1 being the worst and 5 being the best. Normalisation is taken in LLM-as-a-Judge experimental results to maintain a better visualisation. 

\begin{table*}[!t]
\renewcommand{\arraystretch}{1.2}
\small
\centering
\setlength{\tabcolsep}{1.1mm}{
\begin{tabular}{cccccc|cccc|cccc} 
\specialrule{1.5pt}{0pt}{0pt}
\hline
\multirow{2}{*}{\textbf{Backbone}}       & \multirow{2}{*}{\textbf{Methods}} & \multicolumn{4}{c}{\textbf{CC}}                                  & \multicolumn{4}{c}{\textbf{MSC}}                                 & \multicolumn{4}{c}{\textbf{GC}}                                   \\ 
\cline{3-14}
                                         &                                   & \textbf{B-4}  & \textbf{R-L}   & \textbf{Bert}  & \textbf{Mauve} & \textbf{B-4}  & \textbf{R-L}   & \textbf{Bert}  & \textbf{Mauve} & \textbf{B-4}  & \textbf{R-L}   & \textbf{Bert}  & \textbf{Mauve}  \\ 
\hline
\multirow{15}{*}{\textbf{GPT-4o}}       
                                         & MemoChat~\citeyearpar{lu2023memochat}                 & 0.72 & 12.56 & 45.78 & 35.60 & 0.83 & 12.63 & 47.93 & 53.20 & 0.74 & 10.94 & 35.03 & 22.51  \\
                                        & Rsum~\citeyearpar{wang2025recursively}                     & 1.01 & 14.57 & 47.12 & 46.51 & 0.97 & 13.99 & 48.43 & 51.99 & 1.06 & 16.16 & 35.77 & 27.48  \\
                                        & COMEDY~\citeyearpar{chen2025compress}                   & 0.67 & 11.30 & 46.18 & 39.51 & 0.60 & 11.07 & 47.19 & 48.86 & 0.51 & 9.91  & 34.00 & 24.80  \\
                                         
                                         & \multicolumn{13}{c}{{\cellcolor{Iron}}\textit{\textbf{Llama3-Instruct}}}                                                                                                                                                           \\
                                         & \textbf{\textsc{Icml}-1B}                           & 2.31          & 18.72          & 47.62          & 56.63          & 1.42          & 15.30          & 47.99          & 54.71          & 1.21          & 11.09          & 40.74          & 34.39           \\
                                         & \textbf{\textsc{Icml}-3B}                           & 2.37          & 18.78          & 47.65          & \textbf{61.66} & \textbf{1.49} & 15.38          & 48.02          & 57.39          & 1.20          & 11.21          & 40.80          & \uline{36.37}   \\
                                         & \textbf{\textsc{Icml}-8B}                           & 2.31          & 18.29          & 47.40          & 57.76          & \uline{1.46}  & 15.41          & 48.04          & \uline{57.53}  & \textbf{1.25} & 11.26          & 40.84          & 36.42           \\
                                         & \multicolumn{13}{c}{{\cellcolor{Iron}}\textit{\textbf{Gemma3-it}}}                                                                                                                                                                 \\
                                         & \textbf{\textsc{Icml}-1B}                           & 2.25          & 18.88          & \uline{47.76}  & 57.59          & 1.41          & 15.36          & 47.92          & 56.37          & 1.00          & 9.39           & 39.80          & 31.15           \\
                                         & \textbf{\textsc{Icml}-4B}                           & \textbf{2.44} & 18.88          & 47.70          & 58.39          & 1.36          & \uline{15.44}  & 48.01          & 57.02          & 1.16          & \uline{11.32}  & \textbf{40.87} & 35.19           \\
                                         & \textbf{\textsc{Icml}-12B}                          & 2.36          & 18.37          & 47.64          & \uline{58.77}  & 1.44          & 15.27          & 47.97          & 54.46          & 1.19          & 11.25          & 40.75          & 36.13           \\
                                         & \multicolumn{13}{c}{{\cellcolor{Iron}}\textit{\textbf{Qwen3}}}                                                                                                                                                                     \\
                                         & \textbf{\textsc{Icml}-1.7B}                         & 2.19          & \uline{18.92}  & 47.66          & 57.95          & 1.40          & 15.41          & 48.00          & 56.55          & \uline{1.23}  & 11.27          & 40.75          & \textbf{36.69}  \\
                                         & \textbf{\textsc{Icml}-4B}                           & 2.33          & 18.55          & 47.69          & 57.34          & 1.43          & 15.38          & 47.87          & 56.19          & 1.20          & 11.29          & 40.81          & 35.90           \\
                                         & \textbf{\textsc{Icml}-8B}                           & \uline{2.40}  & \textbf{18.93} & 47.74          & 57.60          & 1.40          & \textbf{15.45} & 47.96          & \textbf{57.58} & 1.17          & 11.25          & \uline{40.86}  & 35.85           \\ 
\hline
\multirow{15}{*}{\textbf{Gemini2.5}} 
                                         & MemoChat~\citeyearpar{lu2023memochat}                 & 1.57 & 17.50 & 47.50 & 72.04 & 0.89 & 13.60 & 47.59 & 55.61 & 0.78 & 10.05 & 35.76 & 25.44  \\
                                        & Rsum~\citeyearpar{wang2025recursively}                     & 1.56 & 16.97 & 48.17 & 63.41 & 1.09 & 14.43 & 47.67 & 52.24 & 0.61 & \uline{10.55} & 35.27 & 27.15  \\
                                        & COMEDY~\citeyearpar{chen2025compress}                   & 1.55 & 16.63 & 46.71 & 57.01 & 0.93 & 11.89 & 55.39 & 46.73 & 0.67 & 10.22  & 33.57 & 25.09  \\
                                         
                                         & \multicolumn{13}{c}{{\cellcolor{Iron}}\textit{\textbf{Llama3-Instruct}}}                                                                                                                                                           \\
                                         & \textbf{\textsc{Icml}-1B}                           & 1.94          & 14.48          & 43.15          & 63.95          & 0.94          & 10.83          & 42.56          & 55.59          & 0.92          & 9.05           & 39.78          & 47.38           \\
                                         & \textbf{\textsc{Icml}-3B}                           & \textbf{2.47} & \textbf{18.97} & 47.64          & 78.45          & 1.12          & 13.78          & 46.13          & 64.23          & 0.90          & 9.58           & 39.79          & 43.69           \\
                                         & \textbf{\textsc{Icml}-8B}                           & \textbf{2.47} & 18.37          & 47.36          & 78.62          & \textbf{1.20} & 13.78          & 46.18          & 65.64          & 0.92          & 9.73           & \textbf{40.38}  & 50.42   \\
                                         & \multicolumn{13}{c}{{\cellcolor{Iron}}\textit{\textbf{Gemma3-it}}}                                                                                                                                                                 \\
                                         & \textbf{\textsc{Icml}-1B}                           & 1.74          & 14.47          & 43.56          & 66.39          & 1.11          & 13.65          & 46.00          & \uline{66.15}  & 0.65          & 8.12           & 38.99          & 43.13           \\
                                         & \textbf{\textsc{Icml}-4B}                           & 2.30          & 18.28          & 47.22          & 77.94          & \uline{1.13}  & \uline{13.85}  & 46.20          & \textbf{66.86} & 0.88          & 9.02           & 39.26          & 52.27  \\
                                         & \textbf{\textsc{Icml}-12B}                          & 2.39          & 17.94          & \textbf{47.85} & \uline{78.67}  & 1.07          & 13.77          & 46.12          & 65.89          & 0.77          & 8.37           & 39.40          & 42.10           \\
                                         & \multicolumn{13}{c}{{\cellcolor{Iron}}\textit{\textbf{Qwen3}}}                                                                                                                                                                     \\
                                         & \textbf{\textsc{Icml}-1.7B}                         & 1.98          & 17.25          & 46.15          & 75.48          & 1.07          & 13.56          & 42.55          & 65.50          & \uline{0.95}  & 9.96           & 39.99          & 54.38           \\
                                         & \textbf{\textsc{Icml}-4B}                           & \uline{2.42} & \uline{18.52}  & 47.70          & 77.60          & 1.05          & 13.33          & 46.29          & 65.68          & 0.89          & 10.22           & 39.89          & \uline{57.58}           \\
                                         & \textbf{\textsc{Icml}-8B}                           & 2.21          & 18.22          & \uline{47.82}  & \textbf{80.33} & \uline{1.13}  & \textbf{13.97} & \textbf{48.60} & 66.01          & \textbf{1.28} & \textbf{10.64}  & \uline{40.20} & \textbf{59.81}           \\
\hline
\specialrule{1.5pt}{0pt}{0pt}
\end{tabular}}
\caption{Automatic evaluation (\%) of generation performance per episode. "\textbf{Bold Font}" means the highest results, while "\uline{Underlined Font}" means second-highest results.
*B-4 = BLEU-4, R-L = ROUGE-L, and Bert = BertScore.}
\label{auto_appendix}
\end{table*}

\subsection{Human Metrics}
To further assess the winning performance of different methods in terms of response generation and memory retrieval, we conduct a human evaluation. Following \citet{xu2022long} and \citet{jang2023conversation}, we hire five in-house evaluators to examine 50 randomly selected samples from each of the three datasets. Each sample consists of the model-generated response and its corresponding retrieved memories. The evaluators are tasked with scoring these outputs based on the previously defined metrics to determine which method demonstrates superior capabilities in sustaining high-quality, long-term interactions.

\begin{figure}[!t]
\centering
\includegraphics[width=\linewidth]{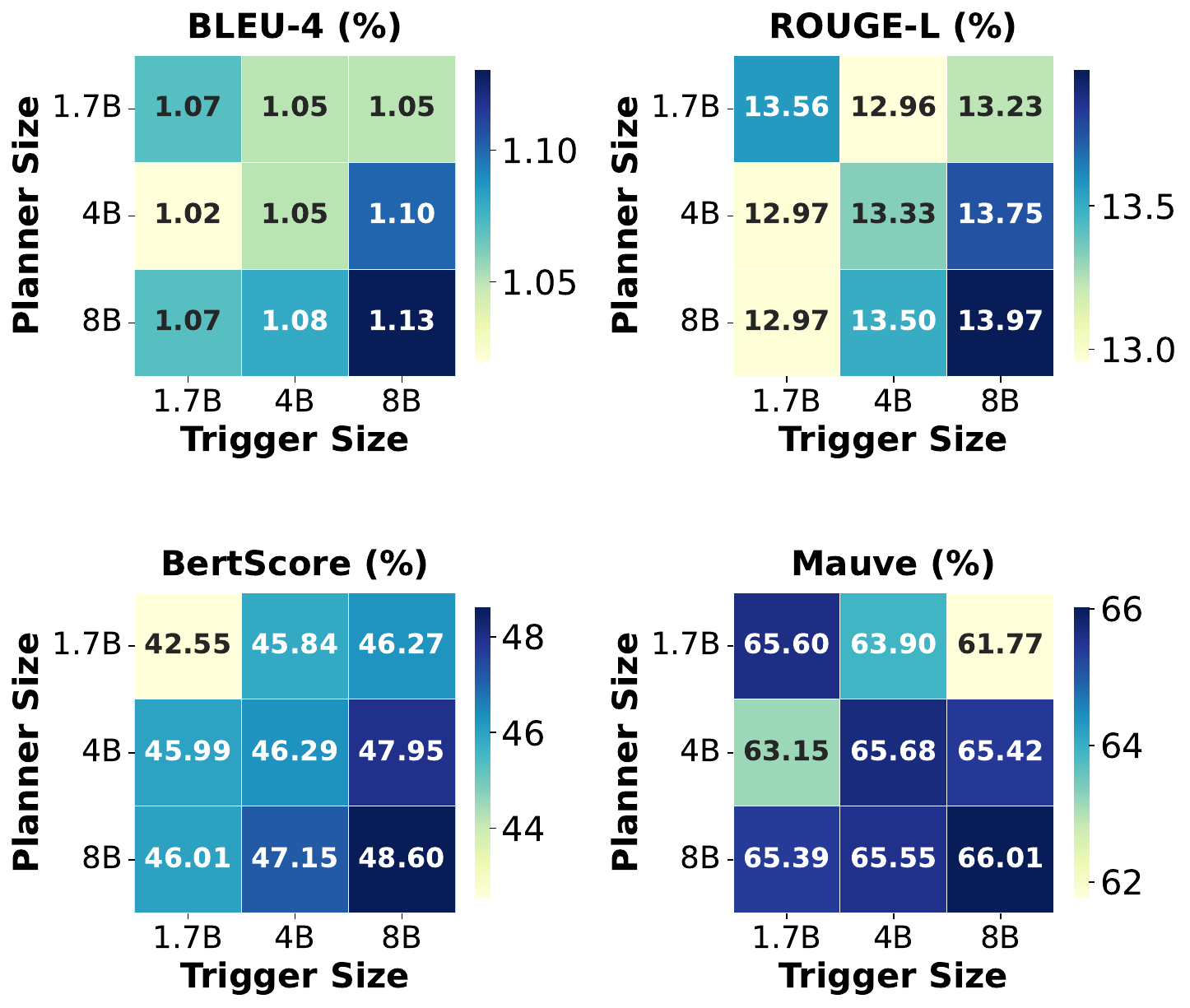}
\caption{Performance scaling of \texttt{Planner} and \texttt{Trigger} on MSC dataset.}
\label{scaling_msc}
\end{figure}

\begin{figure}[!t]
\centering
\includegraphics[width=\linewidth]{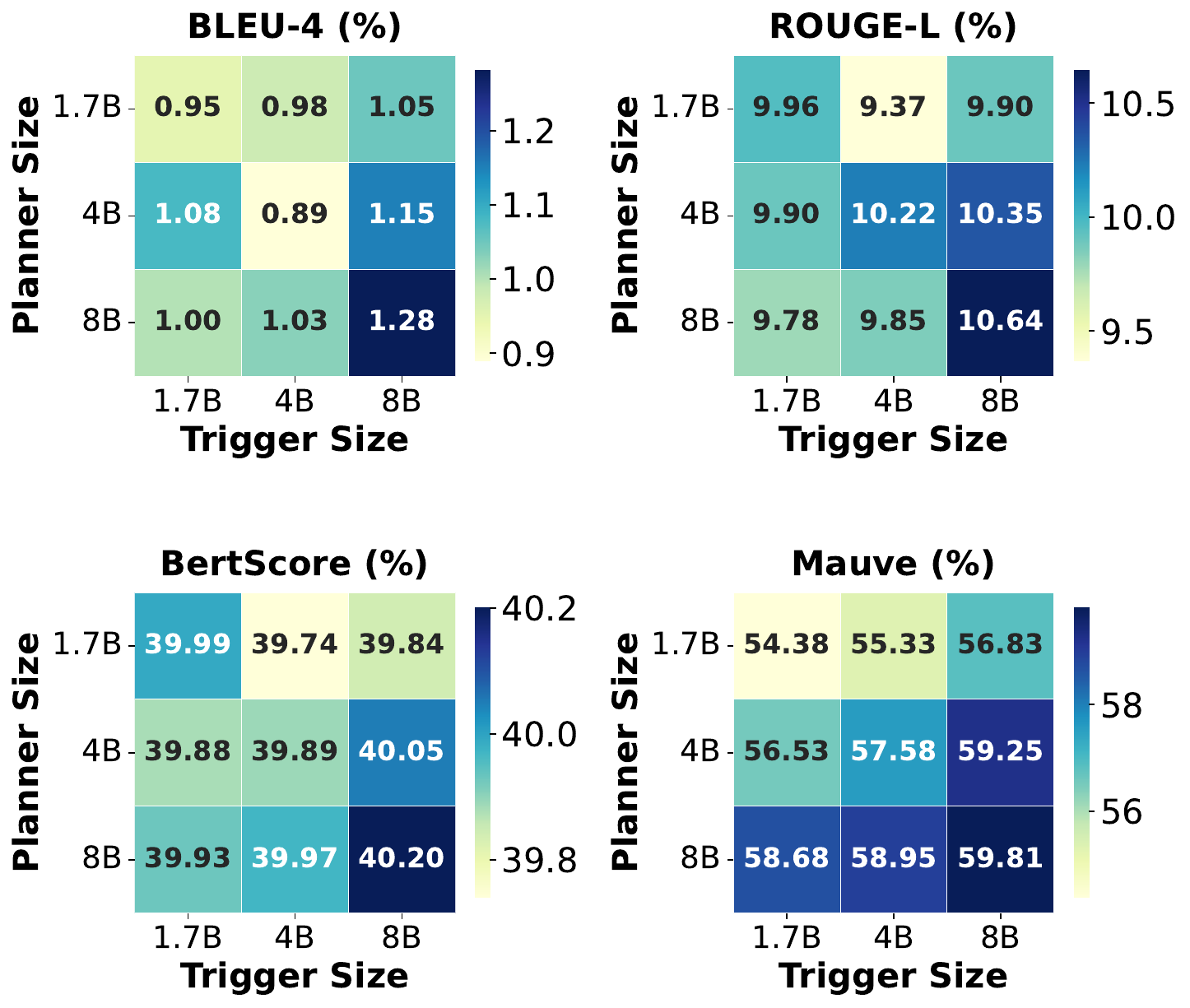}
\caption{Performance scaling of \texttt{Planner} and \texttt{Trigger} on GC dataset.}
\label{scaling_gc}
\end{figure}

\section{Algorithm: Interactive Memory Learning}
\label{algorithm}
This section presents the algorithmic implementation of \textbf{\textsc{Icml}}, illustrating the flow from initialization to online self-evolution. Algorithm \ref{alg:icml} summarizes the complete execution process. The framework begins with supervised pre-training using synthesized expert data $\mathcal{D}_{expert}$ to establish a baseline policy. During real-time interaction, the \texttt{Planner} and \texttt{Trigger} perform on-policy exploration while receiving environmental feedback. A key novelty is the \textbf{Cross-Session Truth Reward} mechanism, which retrospectively aligns planning decisions with the actual conversational utility observed in future sessions.

\section{Performance scaling of \texttt{Planner} and \texttt{Trigger}}
\label{scaling_msc_gc}
Figures~\ref{scaling_msc} and ~\ref{scaling_gc} present the performance scaling results on the MSC and GC datasets. The observations are highly consistent with the findings in our main experiments. Specifically, performance across all metrics generally improves as the sizes of the Planner and Trigger increase, demonstrating a clear scaling effect. Moreover, the diagonal patterns remain evident in these datasets, where matched model sizes often lead to better synergy and more efficient collaboration. These results further confirm that maintaining an architectural balance is a robust strategy for maximizing performance across different data contexts.

\section{Training Reward Analysis}
\label{training_reward}
\begin{figure}[!t]
  \centering
  
  \begin{subfigure}{\linewidth}
    \centering
    \includegraphics[width=\linewidth]{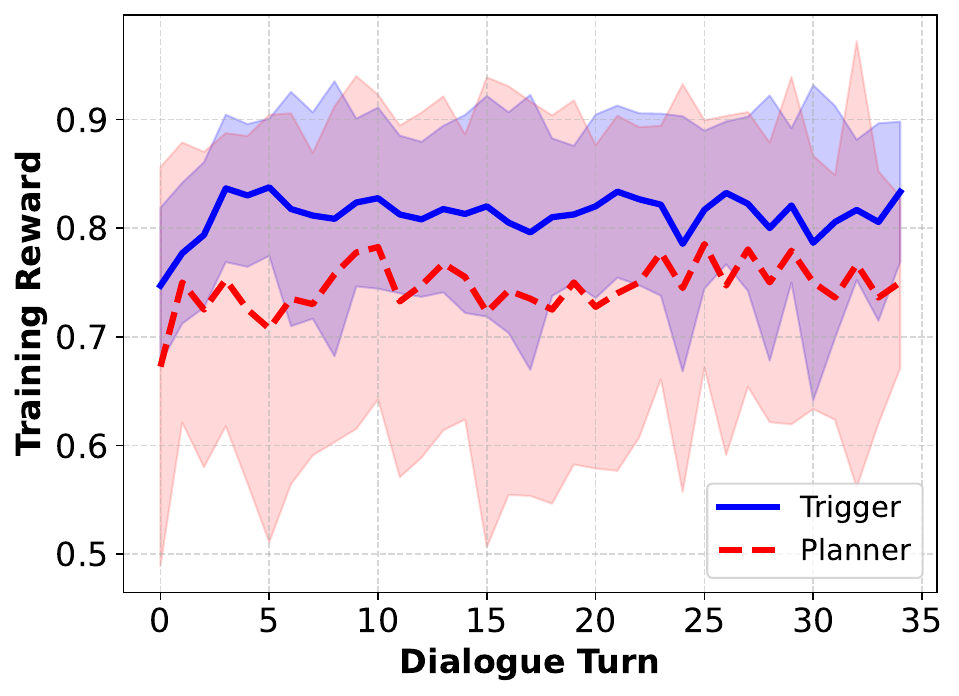}
    \caption{Training reward on CC dataset.}
    \label{fig:reward_cc}
  \end{subfigure}
  \\

  \begin{subfigure}{\linewidth}
    \centering
    \includegraphics[width=\linewidth]{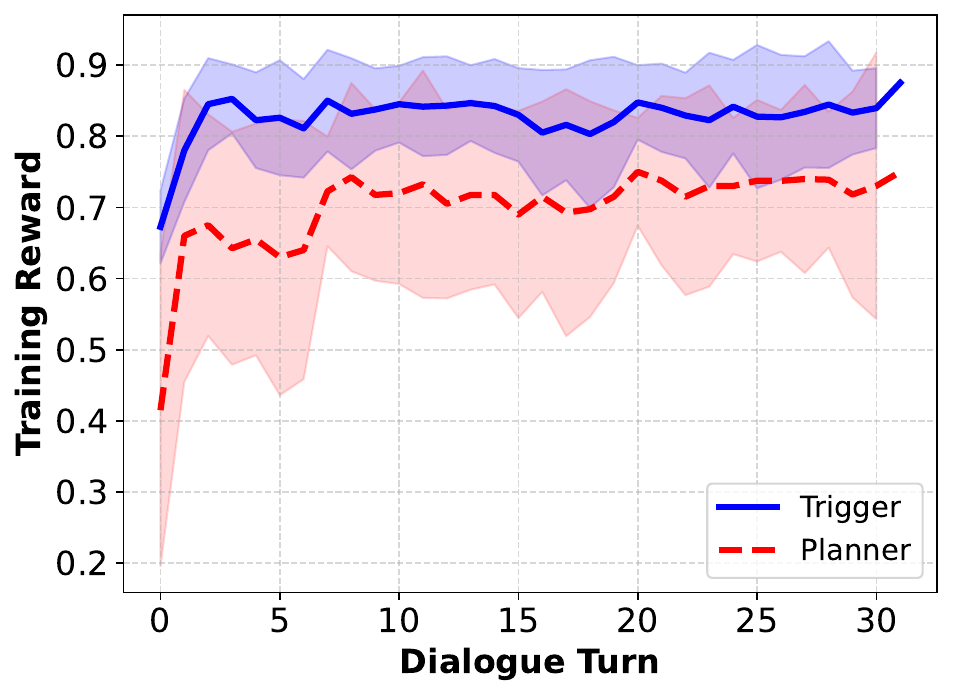}
    \caption{Training reward on MSC dataset.}
    \label{fig:reward_msc}
  \end{subfigure}
  \\

  \begin{subfigure}{\linewidth}
    \centering
    \includegraphics[width=\linewidth]{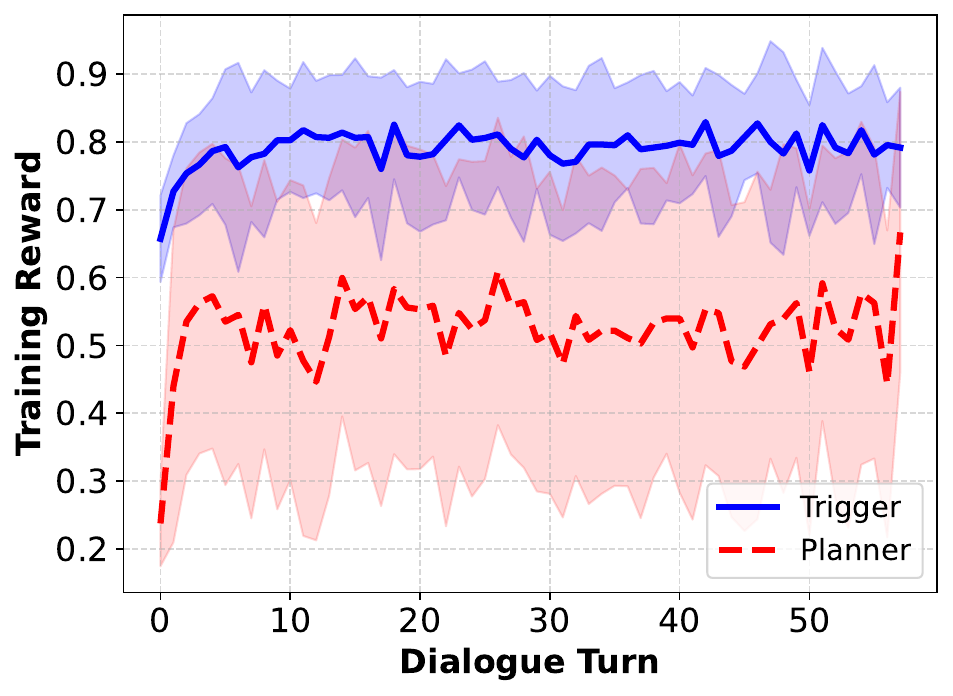}
    \caption{Training reward on GC dataset.}
    \label{fig:reward_gc}
  \end{subfigure}

  \caption{Training reward curves on three different datasets.}
  \label{reward curve}
\end{figure}

We visualize the training reward trajectories for both the \texttt{Planner} and \texttt{Trigger} across the CC, MSC, and GC datasets in Figure~\ref{reward curve}. The curves demonstrate a synchronized upward trend, indicating that the writing and reading policies co-evolve effectively rather than competing adversarially. Crucially, after an initial phase of rapid exploration, the rewards for both agents settle into a stable plateau without significant oscillation. This convergence confirms the robustness of our collaborative reinforcement learning framework, verifying that the system successfully reaches a steady equilibrium where both agents consistently maximize their mutual conversational utility.

\begin{table}[!t]
\renewcommand{\arraystretch}{1.2}
\small
\centering
\setlength{\tabcolsep}{1mm}{
\centering
\begin{tabular}{cccccc} 
\specialrule{1.5pt}{0pt}{0pt}
\hline
\textbf{Backbone}                   & \textbf{Reward Model} & \textbf{B-4} & \textbf{R-L} & \textbf{Bert} & \textbf{Mauve}  \\ 
\hline
\multirow{3}{*}{\textbf{Gemini2.5}} & \cellcolor{Iron}Gemini2.5             & \cellcolor{Iron}2.21         & \cellcolor{Iron}18.22        & \cellcolor{Iron}47.82         & \cellcolor{Iron}80.33           \\ 
\cdashline{2-6}
                                    & GPT-3.5-turbo         & 2.34         & 18.29        & 47.39         & 80.93           \\
                                    & GPT-4o-mini           & 2.20          & 18.31        & 47.29         & 80.69           \\
\hline
\specialrule{1.5pt}{0pt}{0pt}
\end{tabular}}
\caption{Performance comparison using different LLMs as the reward model on the CC dataset.}
\label{reward_robustness}
\end{table}

\section{Robustness to Reward Model Choice}
\label{Robustness}

We test if our method relies on a specific reward model in Table~\ref{reward_robustness}. The results show that performance remains very stable, whether we use \textbf{Gemini2.5}, \textbf{GPT-3.5-turbo}, or \textbf{GPT-4o-mini} as the judge. The differences in key metrics are negligible (e.g., Mauve stays around 80). This proves that the success of \textbf{\textsc{Icml}} comes from its collaborative design, not from the power of the reward model. Therefore, our framework is robust and can work effectively even with smaller or cheaper closed open-source LLMs providing feedback.

\section{Bridging the Gap to User Expectations}
\label{expectation}
\begin{figure}[!t]
\centering
\includegraphics[width=\linewidth]{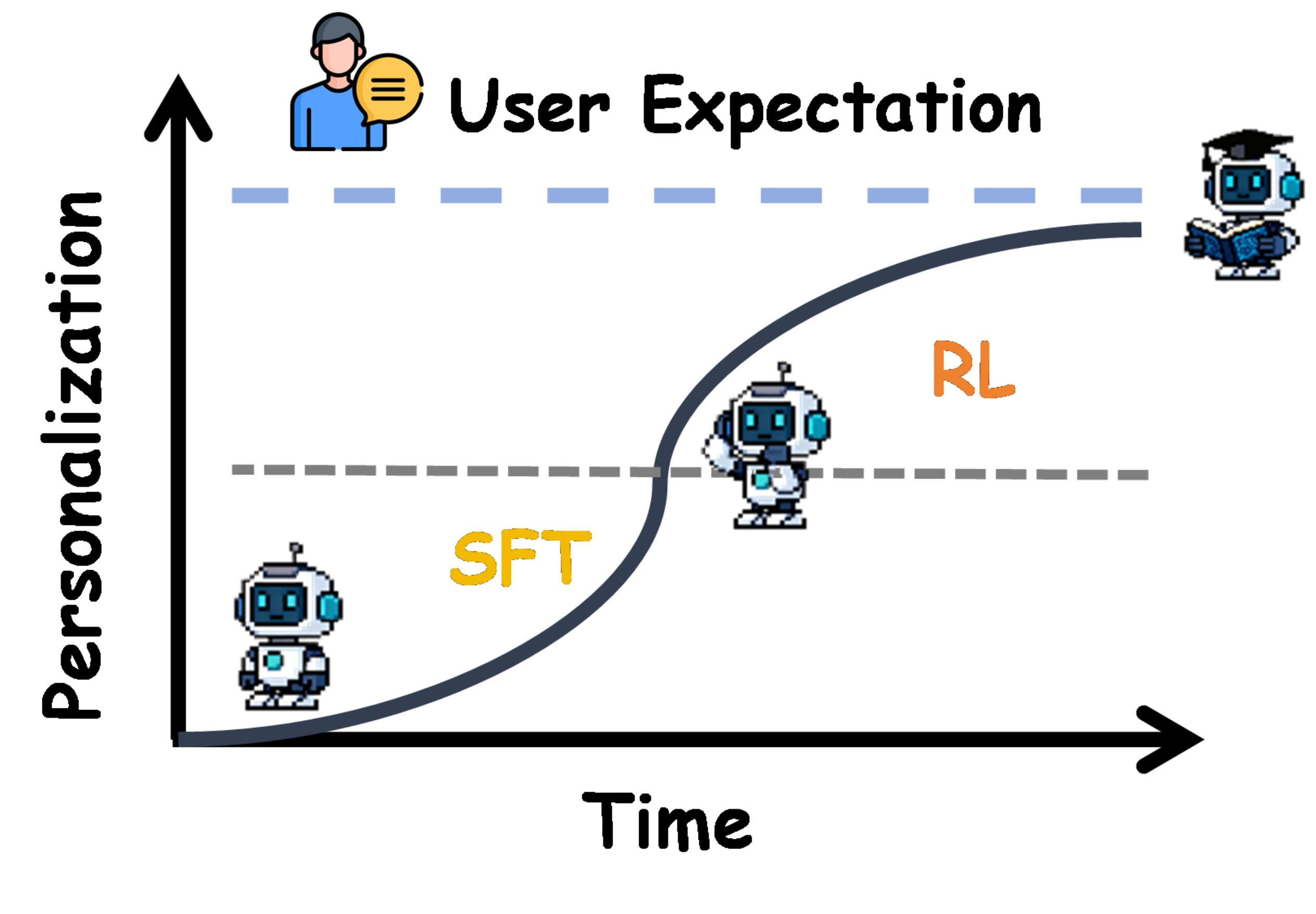}
\caption{Evolution of personalization. SFT means warm-up stage.}
\label{intro_extra}
\end{figure}

Figure~\ref{intro_extra} illustrates the core goal of our framework. Our proposed retrospective synthesis method acts as a crucial "warm-up" stage (labeled as SFT), giving the agent basic memory skills. However, as shown by the grey dashed line, relying only on static synthetic data inevitably hits a performance ceiling because fixed datasets cannot capture constantly changing user behaviors. To break this limit, \textbf{\textsc{Icml}} introduces the reinforcement learning phase. Here, the system treats every real-time interaction as a chance to learn. By using feedback from the environment, the agent actively evolves beyond the static baseline, climbing the curve to finally reach the high level of personalization that users expect.

\section{Efficiency Analysis}
\label{session_efficiency}

\begin{figure}[!t]
\centering
\includegraphics[width=\linewidth]{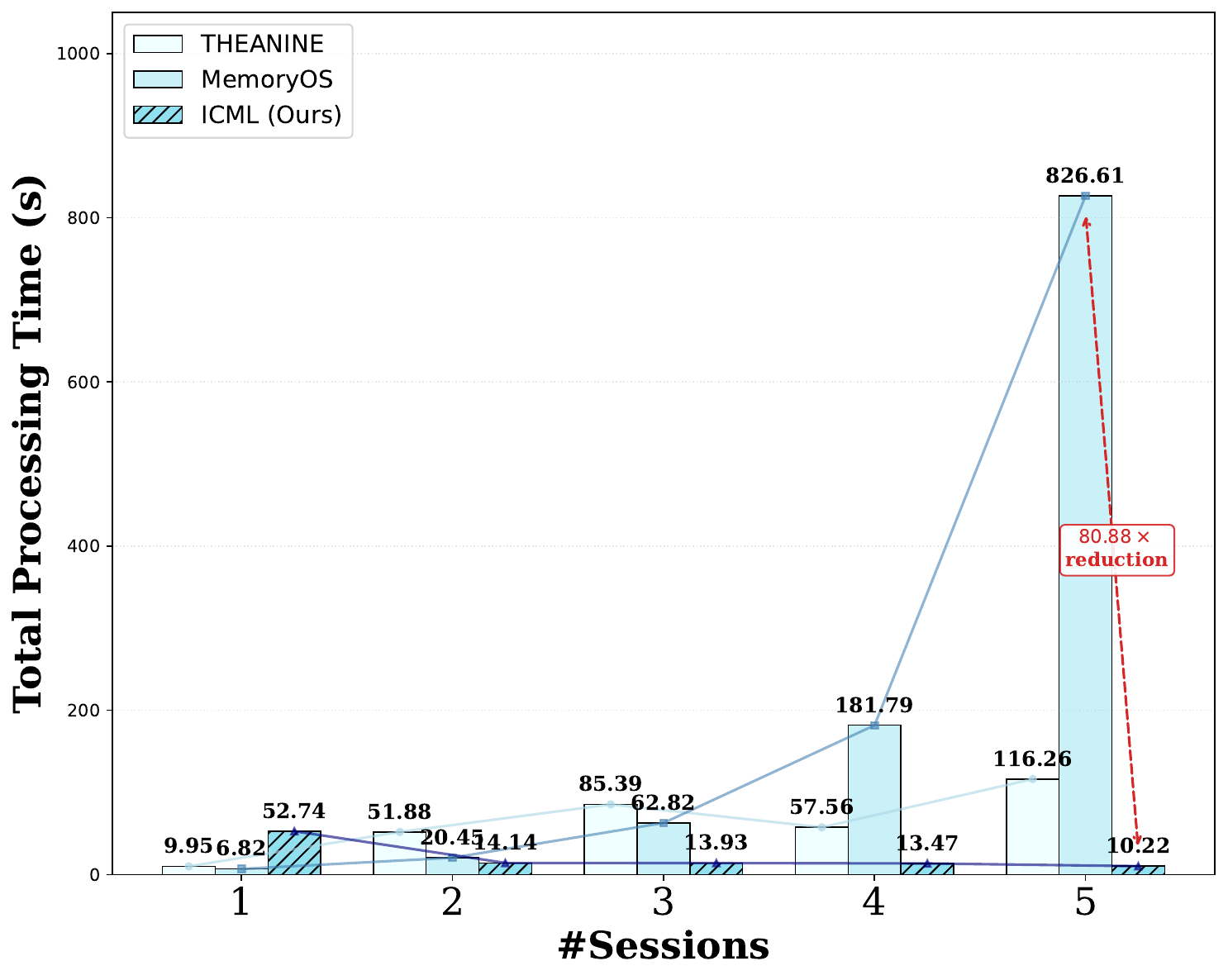}
\caption{Total time efficiency comparison. This demonstrates that our method has sufficient time to adapt to real-world scenarios.}
\label{session_time}
\end{figure}

As illustrated in Figure~\ref{session_time}, our method demonstrates superior long-term efficiency compared to THEANINE and MemoryOS. Although \textbf{\textsc{Icml}} starts with a higher initial cost of 52.74s in Session 1 due to the system's warm-up process—which includes GPU memory allocation and model initialization—it quickly stabilizes to only 10.22s by Session 5. This high-speed stable performance ensures that in real-world interaction scenarios, our method can perform rapid memory updates to support real-time online deployment without significant latency. In contrast, the baselines show significant time increases: THEANINE's cost grows as the memory graph expands, requiring more LLM calls for node relationship checks, while MemoryOS suffers a dramatic time explosion (reaching 826.61s) in late sessions when memory heat triggers heavy long-term memory extraction tasks. Consequently, ICML effectively maintains a constant and efficient processing speed for long conversations, avoiding the performance bottlenecks found in traditional graph-based or hierarchical memory systems.

\section{Analysis of Early Exploration and Proxy Rewards}
\label{sec:appendix_exploration}

In the early stages of interactive learning, the Trigger agent may occasionally fail to retrieve useful memories due to insufficient exploration. This raises a potential concern regarding exploration failure: if a valuable memory is stored but never retrieved, it will not receive the delayed Cross-Session Truth Reward, which might seemingly hinder the Planner agent's ability to learn. 

To address this issue, our framework does not rely solely on the delayed Truth Reward. As detailed in Section 3.3.1, we incorporate an immediate Proxy Reward ($r^{proxy}$) and a Miss-penalty ($-\alpha r^{proxy}$) provided by the backbone LLM. Even if the Trigger fails to select a memory later, the Planner receives immediate feedback regarding the intrinsic value of the dialogue turn. This mechanism ensures that high-value information is consistently retained during the initial exploration phase.

To empirically validate the effectiveness of this mechanism, we tracked the memory retention performance and reward dynamics over five continuous training sessions. We measured the Miss Rate of high-value memories, the False Positive Rate (FPR), and the Pearson correlation ($r$) between the immediate Proxy Reward and the delayed Truth Reward. 

\begin{table*}[!t]
\centering
\small
\begin{tabular}{lccc}
\specialrule{1.5pt}{0pt}{0pt}
\toprule
\textbf{Training Session} & \textbf{Miss Rate (\%)} & \textbf{False Positive Rate (\%)} & \textbf{Proxy \& Truth Reward Correlation ($r$)} \\
\midrule
Session 1 (Start) & 48.2 & 15.6 & 0.62 \\
Session 2         & 24.5 & 12.1 & 0.75 \\
Session 3         & 5.8  & 8.4  & 0.83 \\
Session 4         & 4.1  & 7.2  & 0.87 \\
Session 5         & 3.2  & 6.1  & 0.89 \\
\specialrule{1.5pt}{0pt}{0pt}
\bottomrule
\end{tabular}
\caption{Performance metrics of memory exploration over training sessions. The Proxy Reward \& Truth Reward Correlation is measured using Pearson $r$.}
\label{tab:exploration_metrics}
\end{table*}

As shown in Table \ref{tab:exploration_metrics}, there is a clear and rapid downward trend in the Miss Rate from Session 1 to Session 3. This rapid drop confirms that the Proxy Rewards effectively and quickly guide the Planner's exploration before the Truth Rewards become sufficiently dense. Furthermore, the continued improvement in later sessions (Sessions 4 and 5) and the steadily increasing correlation ($r$) demonstrate that the model successfully refines its policy over prolonged interactions, achieving higher precision and stronger alignment between the internal proxy evaluation and the actual long-term user expectations.

\section{Significance and Agreement Analysis for Human Evaluation}
\label{sec:appendix_human_eval_stats}

To confirm the reliability and statistical significance of our human evaluation results, we conducted further statistical tests comparing our ICML framework against the strongest baseline, THEANINE. Specifically, we calculated Fleiss' Kappa ($\kappa$) to measure the agreement among our five in-house evaluators, and we computed the 95\% Confidence Intervals along with P-values (using pairwise t-tests) for the win rates.

\begin{table*}[!t]
\small
\centering
\begin{tabular}{lcccc}
\specialrule{1.5pt}{0pt}{0pt}
\toprule
\textbf{Metric} & \textbf{ICML Win Rate (\%)} & \textbf{95\% Confidence Interval} & \textbf{Fleiss' Kappa ($\kappa$)} & \textbf{P-value (vs. Baseline)} \\
\midrule
Generation Quality & 66.0 & [60.5\%, 71.5\%] & 0.65 (Substantial) & $< 0.01$ \\
Memory Retrieval   & 70.0 & [64.2\%, 75.8\%] & 0.68 (Substantial) & $< 0.01$ \\
\specialrule{1.5pt}{0pt}{0pt}
\bottomrule
\end{tabular}
\caption{Human Evaluation Statistics (ICML vs. THEANINE).}
\label{tab:human_eval_stats}
\end{table*}

As shown in Table \ref{tab:human_eval_stats}, the Fleiss' Kappa scores ($>0.6$) demonstrate a reliable, substantial agreement among the evaluators. Furthermore, the win rates of 66.0\% and 70.0\% are statistically significant ($p < 0.01$) and feature narrow confidence intervals. This rigorously validates that the observed superiority of ICML in both generation quality and memory retrieval is highly significant and not attributable to random chance or small sample size noise.

\section{Hyperparameter Sensitivity Analysis}
\label{sec:appendix_sensitivity}

To evaluate the robustness of our hybrid reward design, we conducted comprehensive sensitivity experiments regarding the hyperparameter $\lambda$, which balances the Immediate Proxy Reward and the Delayed Quality Reward. Using ICML-8B (Qwen3), we tested $\lambda \in \{0.3, 0.5, 0.7\}$ across the CC, MSC, and GC datasets, evaluated by two different backbone models (GPT-4o and Gemini2.5). 

\begin{table*}[!t]
\centering
\small
\begin{tabular}{lllcccc}
\specialrule{1.5pt}{0pt}{0pt}
\toprule
\textbf{Backbone} & \textbf{Dataset} & \textbf{$\lambda$ Value} & \textbf{B-4} & \textbf{R-L} & \textbf{Bert} & \textbf{Mauve} \\
\midrule
\multirow{9}{*}{GPT-4o} & \multirow{3}{*}{CC} & $\lambda = 0.3$ & 2.31 & 18.45 & 47.12 & 55.80 \\
& & $\lambda = 0.5$ (Main) & \textbf{2.40} & \textbf{18.93} & \textbf{47.74} & \textbf{57.60} \\
& & $\lambda = 0.7$ & 2.35 & 18.62 & 47.35 & 56.45 \\
\cmidrule{2-7}
& \multirow{3}{*}{MSC} & $\lambda = 0.3$ & 1.28 & 14.85 & 47.15 & 56.13 \\
& & $\lambda = 0.5$ (Main) & \textbf{1.40} & \textbf{15.45} & \textbf{47.96} & \textbf{57.58} \\
& & $\lambda = 0.7$ & 1.35 & 15.19 & 47.54 & 56.91 \\
\cmidrule{2-7}
& \multirow{3}{*}{GC} & $\lambda = 0.3$ & 1.12 & 10.95 & 40.15 & 34.50 \\
& & $\lambda = 0.5$ (Main) & \textbf{1.17} & \textbf{11.25} & \textbf{40.86} & \textbf{35.85} \\
& & $\lambda = 0.7$ & 1.15 & 11.10 & 40.57 & 35.14 \\
\midrule
\multirow{9}{*}{Gemini2.5} & \multirow{3}{*}{CC} & $\lambda = 0.3$ & 2.15 & 17.82 & 47.26 & 78.54 \\
& & $\lambda = 0.5$ (Main) & \textbf{2.21} & \textbf{18.22} & \textbf{47.82} & \textbf{80.33} \\
& & $\lambda = 0.7$ & 2.18 & 18.05 & 47.55 & 79.10 \\
\cmidrule{2-7}
& \multirow{3}{*}{MSC} & $\lambda = 0.3$ & 1.08 & 13.51 & 48.12 & 64.20 \\
& & $\lambda = 0.5$ (Main) & \textbf{1.13} & \textbf{13.97} & \textbf{48.60} & \textbf{66.01} \\
& & $\lambda = 0.7$ & 1.10 & 13.75 & 48.35 & 65.10 \\
\cmidrule{2-7}
& \multirow{3}{*}{GC} & $\lambda = 0.3$ & 1.22 & 10.15 & 39.87 & 58.23 \\
& & $\lambda = 0.5$ (Main) & \textbf{1.28} & \textbf{10.64} & \textbf{40.20} & \textbf{59.81} \\
& & $\lambda = 0.7$ & 1.25 & 10.44 & 40.05 & 58.97 \\
\specialrule{1.5pt}{0pt}{0pt}
\bottomrule
\end{tabular}
\caption{Sensitivity analysis of the hyperparameter $\lambda$ using ICML-8B (Qwen3) across different backbones and datasets.}
\label{tab:sensitivity_lambda}
\end{table*}

As shown in Table \ref{tab:sensitivity_lambda}, the setting of $\lambda = 0.5$ (used in our main experiments) consistently yields the best performance across both backbones and all three datasets. Minor deviations ($\lambda = 0.3$ or $0.7$) result in slight performance drops. This confirms that while $\lambda$ influences the trade-off between short-term guidance and long-term objectives, our chosen hyperparameter is optimal, and the system remains stable without drastic collapse regardless of the evaluator architecture.

\section{Robustness to Tiny Reward Models}
\label{sec:appendix_tiny_reward}

To stress-test the robustness of our framework, we conducted additional experiments employing significantly weaker LLMs (Qwen2.5-1.5B-Instruct and Llama-3.2-1B-Instruct) as Reward Models (Judges). We evaluated the impact on performance across the CC, MSC, and GC datasets using ICML-8B (Qwen3). 

\begin{table*}[!t]
\centering
\small
\begin{tabular}{lllcccc}
\specialrule{1.5pt}{0pt}{0pt}
\toprule
\textbf{Backbone} & \textbf{Dataset} & \textbf{Reward Model (Judge)} & \textbf{B-4 (\%)} & \textbf{R-L (\%)} & \textbf{Bert (\%)} & \textbf{Mauve (\%)} \\
\midrule
\multirow{9}{*}{GPT-4o} & \multirow{3}{*}{CC} & GPT-4o (Baseline) & \textbf{2.40} & \textbf{18.93} & \textbf{47.74} & \textbf{57.60} \\
& & Qwen2.5-1.5B & 2.25 & 18.14 & 47.05 & 54.82 \\
& & Llama-3.2-1B & 2.18 & 17.85 & 46.86 & 53.20 \\
\cmidrule{2-7}
& \multirow{3}{*}{MSC} & GPT-4o (Baseline) & \textbf{1.40} & \textbf{15.45} & \textbf{47.96} & \textbf{57.58} \\
& & Qwen2.5-1.5B & 1.31 & 14.83 & 47.24 & 55.10 \\
& & Llama-3.2-1B & 1.25 & 14.57 & 46.92 & 53.85 \\
\cmidrule{2-7}
& \multirow{3}{*}{GC} & GPT-4o (Baseline) & \textbf{1.17} & \textbf{11.25} & \textbf{40.86} & \textbf{35.85} \\
& & Qwen2.5-1.5B & 1.09 & 10.81 & 40.19 & 33.92 \\
& & Llama-3.2-1B & 1.05 & 10.54 & 39.80 & 32.59 \\
\midrule
\multirow{9}{*}{Gemini2.5} & \multirow{3}{*}{CC} & Gemini2.5 (Baseline) & \textbf{2.21} & \textbf{18.22} & \textbf{47.82} & \textbf{80.33} \\
& & Qwen2.5-1.5B & 2.08 & 17.54 & 47.12 & 76.50 \\
& & Llama-3.2-1B & 1.98 & 17.18 & 46.85 & 74.27 \\
\cmidrule{2-7}
& \multirow{3}{*}{MSC} & Gemini2.5 (Baseline) & \textbf{1.13} & \textbf{13.97} & \textbf{48.60} & \textbf{66.01} \\
& & Qwen2.5-1.5B & 1.05 & 13.42 & 47.90 & 63.52 \\
& & Llama-3.2-1B & 1.01 & 13.11 & 47.56 & 61.83 \\
\cmidrule{2-7}
& \multirow{3}{*}{GC} & Gemini2.5 (Baseline) & \textbf{1.28} & \textbf{10.64} & \textbf{40.20} & \textbf{59.81} \\
& & Qwen2.5-1.5B & 1.18 & 10.18 & 39.53 & 56.44 \\
& & Llama-3.2-1B & 1.12 & 9.88 & 39.12 & 54.52 \\
\specialrule{1.5pt}{0pt}{0pt}
\bottomrule
\end{tabular}
\caption{Performance comparison of ICML-8B (Qwen3) when trained with tiny Reward Models (Judges).}
\label{tab:tiny_reward_models}
\end{table*}

As shown in Table \ref{tab:tiny_reward_models}, even when guided by a 1B-parameter judge, the performance degradation is minimal compared to the strong teacher models (GPT-4o and Gemini2.5). While there is an expected slight drop in metrics, the system does not collapse and continues to perform robustly. This highlights the resilience of the ICML framework to noisy reward signals and demonstrates its strong potential for deployment in resource-constrained environments.

\section{Effectiveness of Backward Storyline Generation}
\label{sec:appendix_backward_vs_forward}

To clearly demonstrate the motivation behind our Backward Storyline Generation, we explicitly compared it against standard Forward Generation. By anchoring the generation on the future outcome (the Seed Session) and generating backwards, we force the LLM to plant necessary clues that causally lead to the current outcome (e.g., ensuring the generated history logically explains the user's current constraints). In contrast, forward generation lacks this target-driven control and often fails to converge to the specific constraints required for effective cold-start training.

To empirically validate this, we evaluated the performance of ICML-8B (based on Qwen3) trained with initialization data generated from both strategies. We tested across three datasets using GPT-4o and Gemini2.5 as backbones.

\begin{table*}[!t]
\centering
\small
\begin{tabular}{lllccccc}
\specialrule{1.5pt}{0pt}{0pt}
\toprule
\textbf{Dataset} & \textbf{Backbone} & \textbf{Method} & \textbf{B-4} & \textbf{R-L} & \textbf{Bert} & \textbf{Mauve} & \textbf{LLM Judge (1-5)} \\
\midrule
\multirow{4}{*}{CC} & \multirow{2}{*}{GPT-4o} & Forward & 1.85 & 16.51 & 45.20 & 52.15 & 4.15 \\
& & Backward (Ours) & \textbf{2.40} & \textbf{18.93} & \textbf{47.74} & \textbf{57.60} & \textbf{4.82} \\
\cmidrule{2-8}
& \multirow{2}{*}{Gemini2.5} & Forward & 1.92 & 16.85 & 45.82 & 70.24 & 4.22 \\
& & Backward (Ours) & \textbf{2.21} & \textbf{18.22} & \textbf{47.82} & \textbf{80.33} & \textbf{4.88} \\
\midrule
\multirow{4}{*}{MSC} & \multirow{2}{*}{GPT-4o} & Forward & 1.15 & 13.20 & 46.15 & 51.58 & 4.08 \\
& & Backward (Ours) & \textbf{1.40} & \textbf{15.45} & \textbf{47.96} & \textbf{57.58} & \textbf{4.76} \\
\cmidrule{2-8}
& \multirow{2}{*}{Gemini2.5} & Forward & 0.95 & 12.15 & 46.50 & 60.55 & 4.12 \\
& & Backward (Ours) & \textbf{1.13} & \textbf{13.97} & \textbf{48.60} & \textbf{66.01} & \textbf{4.85} \\
\midrule
\multirow{4}{*}{GC} & \multirow{2}{*}{GPT-4o} & Forward & 1.05 & 9.55 & 38.52 & 32.17 & 3.95 \\
& & Backward (Ours) & \textbf{1.17} & \textbf{11.25} & \textbf{40.86} & \textbf{35.85} & \textbf{4.68} \\
\cmidrule{2-8}
& \multirow{2}{*}{Gemini2.5} & Forward & 1.02 & 9.12 & 38.25 & 52.45 & 4.05 \\
& & Backward (Ours) & \textbf{1.28} & \textbf{10.64} & \textbf{40.20} & \textbf{59.81} & \textbf{4.79} \\
\specialrule{1.5pt}{0pt}{0pt}
\bottomrule
\end{tabular}
\caption{Comparison of ICML-8B (Qwen3) trained with Backward vs. Forward synthetic data strategies.}
\label{tab:backward_vs_forward}
\end{table*}

As shown in Table \ref{tab:backward_vs_forward}, the Backward (Ours) strategy consistently outperforms Forward generation across all metrics. This empirical evidence confirms that backward generation provides a higher-quality, logically consistent initialization signal, which is crucial for the subsequent reinforcement learning stage.

\section{Comparison with RMM}
\label{sec:appendix_rmm_comparison}

To further evaluate the zero-shot online adaptation capability of our framework, we compared ICML against Reflective Memory Management (RMM) \cite{tan2025prospect}. Since RMM also proposes an online reinforcement learning framework to optimize memory management, we implemented a variant of RMM that utilizes only its online RL module (Retrospective Reflection) while skipping the offline supervised pre-training. This ensures a fair comparison under a strict zero-shot test-time adaptation setting, where neither model has access to the task-specific training sets.

We conducted a comprehensive evaluation across the CC, MSC, and GC datasets using GPT-4o and Gemini 2.5 as backbones. Both methods operate in a zero-shot setting regarding the dataset, and our method utilizes ICML-8B (based on Qwen3) as the memory policy model.

\begin{table*}[!t]
\centering
\small
\begin{tabular}{lllcccc}
\specialrule{1.5pt}{0pt}{0pt}
\toprule
\textbf{Dataset} & \textbf{Backbone} & \textbf{Method} & \textbf{B-4} & \textbf{R-L} & \textbf{Bert} & \textbf{Mauve} \\
\midrule
\multirow{4}{*}{CC} & \multirow{2}{*}{GPT-4o} & RMM (RL-only) & 1.95 & 17.10 & 46.54 & 52.39 \\
& & ICML (Ours) & \textbf{2.40} & \textbf{18.93} & \textbf{47.74} & \textbf{57.60} \\
\cmidrule{2-7}
& \multirow{2}{*}{Gemini 2.5} & RMM (RL-only) & 1.85 & 16.92 & 46.25 & 72.42 \\
& & ICML (Ours) & \textbf{2.21} & \textbf{18.22} & \textbf{47.82} & \textbf{80.33} \\
\midrule
\multirow{4}{*}{MSC} & \multirow{2}{*}{GPT-4o} & RMM (RL-only) & 1.20 & 14.27 & 46.84 & 52.62 \\
& & ICML (Ours) & \textbf{1.40} & \textbf{15.45} & \textbf{47.96} & \textbf{57.58} \\
\cmidrule{2-7}
& \multirow{2}{*}{Gemini 2.5} & RMM (RL-only) & 0.98 & 12.59 & 47.61 & 63.47 \\
& & ICML (Ours) & \textbf{1.13} & \textbf{13.97} & \textbf{48.60} & \textbf{66.01} \\
\midrule
\multirow{4}{*}{GC} & \multirow{2}{*}{GPT-4o} & RMM (RL-only) & 1.08 & 10.32 & 39.56 & 33.04 \\
& & ICML (Ours) & \textbf{1.17} & \textbf{11.25} & \textbf{40.86} & \textbf{35.85} \\
\cmidrule{2-7}
& \multirow{2}{*}{Gemini 2.5} & RMM (RL-only) & 1.10 & 9.87 & 39.53 & 54.02 \\
& & ICML (Ours) & \textbf{1.28} & \textbf{10.64} & \textbf{40.20} & \textbf{59.81} \\
\bottomrule
\specialrule{1.5pt}{0pt}{0pt}
\end{tabular}
\caption{Comparison with RMM (RL-only) across different datasets and backbones in a zero-shot setting.}
\label{tab:rmm_comparison}
\end{table*}

As shown in Table \ref{tab:rmm_comparison}, ICML consistently outperforms the RL-only variant of RMM across all datasets and backbones. While RMM is a strong baseline, its performance drops significantly when deprived of offline training data. This empirical evidence confirms that ICML's dual-agent architecture is far more effective for zero-shot online adaptation, demonstrating superior data efficiency and adaptability in true zero-shot scenarios compared to methods that heavily rely on offline supervision for initialization.

\section{Retrospective Session Synthesis Prompts}
\label{data_prompt}
The following subsections describe the prompts used in the \textbf{Retrospective Session Synthesis} pipeline, which initializes the \textbf{\textsc{Icml}} framework with high-quality expert trajectories.

\subsection{Backward Storyline Generation}
To address the cold-start problem where agents lack historical context, we utilize a reverse-generation strategy. Starting from a seed session $S_{seed}$, this prompt guides the LLM to recursively generate preceding sessions that provide logical grounding for the user's current preferences or constraints. This ensures that the generated history is both consistent and causally linked to the final interaction. The specific instruction set for this stage is presented in Figure \ref{fig:Prequel Generation}.

\begin{figure*}[!t]
\begin{tcolorbox}[
colframe=black!75!white, 
colback=white, sharp corners, 
boxrule=0.8pt, width=\textwidth,
title=Prompt for Prequel Session Generation
] 
"""\\
\textbf{\# Role}\\
You are a professional screenwriter, skilled at creating natural, logically coherent, and emotionally resonant human-like dialogues.\\
\textbf{\# Background}\\
The two characters are a User and a Chat AI. Their initial interactions are used as Ground Truth. The following dialogues are listed in reverse chronological order.\\
\{storyline\_hint\}\\
\{known\_dialogue\_str\}\\
\textbf{\# Your Task}\\
Please create a new dialogue that occurs before all these known dialogues, serving as their prequel. This new dialogue should lay the foundation or foreshadow topics in the known dialogues, maintain consistent character styles, and try to be as distinct as possible from previous creations. The dialogue should exceed 20 turns.\\
\textbf{\# Format Requirements}\\
Please output strictly in the following JSON format, containing a "dialogue" list with multiple rounds of conversation, without any reasoning.\\
\{\\
\quad "dialogue": [\\
\qquad \{"speaker": "User", "content": "..."\},\\
\qquad \{"speaker": "Bot", "content": "..."\}\\
\quad ]\\
\}\\
"""
\end{tcolorbox} 
\caption{Prompt for backward prequel session synthesis.}
\label{fig:Prequel Generation}
\end{figure*}

\subsection{Forward Dependency Annotation: \texttt{Planner} Agent}
Once the storyline is established, we perform \textit{Forward Dependency Annotation}. This prompt corresponds to the \texttt{Planner}'s role in the synthesis phase. It evaluates the information gain of each dialogue turn to determine whether it contains high-value information worth saving. This process creates the binary labels necessary for the agent to learn how to distinguish critical user facts from transient noise. As illustrated in Figure \ref{fig:Value Evaluation}, the agent is instructed to focus on implicit traits and potential future topics.

\begin{figure*}[!t]
\begin{tcolorbox}[
colframe=black!75!white, 
colback=white, sharp corners, 
boxrule=0.8pt, width=\textwidth,
title=Prompt for Information Value Evaluation.
] 
"""\\
\textbf{\# Role}\\
You are an information value evaluation expert, inclined to capture more potentially useful information.

\textbf{\# Task}\\
Given the current turn and context of the dialogue, determine if the "current turn" contains information that is **possibly** worth remembering in the long term.

\textbf{\# Criteria}\\
In addition to explicit key information (facts, preferences, agreements, important events), please also consider:\\
- Details that hint at character personality, emotions, or motivations.\\
- Topics that may be indirectly mentioned or serve as background in future dialogues.\\
- Fragments that help understand the overall flow of the dialogue, even if they are not core.

\textbf{\# Dialogue Snippet}\\
- Context (previous turn): "\{context\_str\}"\\
- Current turn: "\{current\_turn\_content\}"

\textbf{\# Output Requirements}\\
Please answer strictly with \texttt{true} or \texttt{false}. Please be more inclined to answer \texttt{true}.\\
"""
\end{tcolorbox} 
\caption{Prompt for evaluating the information value of dialogue turns.}
\label{fig:Value Evaluation}
\end{figure*}

\subsection{Forward Dependency Annotation: \texttt{Trigger} Agent}
This prompt facilitates the annotation of memory dependencies for the \texttt{Trigger}. By identifying which specific historical fragments are required to resolve a query in the current session, we establish the explicit links between retrieval actions and conversational utility. This annotation allows the policy to be warmed up with dense, causal signals before online reinforcement learning begins. We utilize the logic detailed in Figure \ref{fig:Memory Linking} to perform this fine-grained association analysis.

\begin{figure*}[!t]
\begin{tcolorbox}[
colframe=black!75!white, 
colback=white, sharp corners, 
boxrule=0.8pt, width=\textwidth,
title=Prompt for Memory Linking Analysis
] 
"""\\
\textbf{\# Role}\\
You are a dialogue logic analysis expert, skilled at identifying explicit and implicit associations in conversations.

\textbf{\# Task}\\
Determine if the "current response" is associated with or potentially references any memory in the "history memory list". Even if it is not a direct quote, as long as the content is related, inspired by, or a continuation of past topics, it should be considered an association.

\textbf{\# History Memory List}\\
\{memory\_list\_str\}

\textbf{\# Current Dialogue}\\
- Query (previous turn): "\{prev\_turn\_content\}"\\
- Current response: "\{current\_turn\_content\}"

\textbf{\# Output Requirements}\\
If an association exists, please return the associated \texttt{memory\_id} list (can be one or more). If there is no obvious association, return \texttt{null}.\\
Please output strictly in JSON format. For example: \{\texttt{"used\_memory\_ids": ["s1\_t5\_mem", "s2\_t1\_mem"]}\} or \{\texttt{"used\_memory\_ids": null}\}.\\
Please be more active in searching for association relationships.\\
"""
\end{tcolorbox} 
\caption{Prompt for linking current turns to historical memory items.}
\label{fig:Memory Linking}
\end{figure*}

\section{Online Interaction and RL Training Prompts}
\label{training_prompt}
This section details the prompts used during the active interaction phase, where the \texttt{Planner} and \texttt{Trigger} agents co-evolve based on environmental feedback.

\subsection{Agent Response Generation}
To produce human-like and personalized replies, the agent's generation process is grounded in both the current dialogue history and the retrieved memory fragments. As described in Figure \ref{fig:Response Generation}, the prompt instructs the model to synthesize this information while maintaining brevity and ignoring misleading context.

\begin{figure*}[!t]
\begin{tcolorbox}[
colframe=black!75!white, 
colback=white, sharp corners, 
boxrule=0.8pt, width=\textwidth,
title=Prompt for Agent Response Generation
] 
"""\\
Relevant Memories: \\
- \{selected\_memory\_content\}\\

Current Conversation: \{user\_utterance\}\\

Generate the most plausible next response like a human based on the current conversation. You can refer to your memory, but you should ignore the memory if it misleads the next response. Do not put too much information in the next response.\\
"""
\end{tcolorbox} 
\caption{Prompt for generating personalized agent responses grounded in memory.}
\label{fig:Response Generation}
\end{figure*}

\subsection{\texttt{Planner} Agent: Proxy Reward Evaluation}
\begin{figure*}[!h]
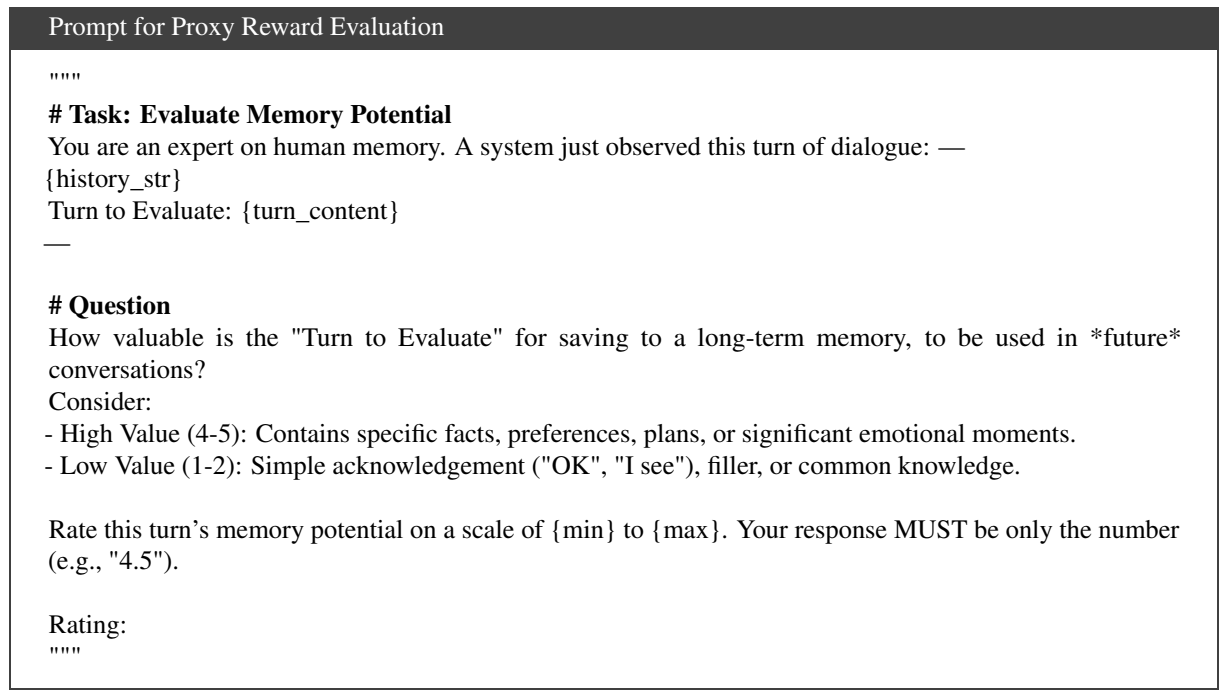

\begin{tcolorbox}[
colframe=black!75!white, 
colback=white, sharp corners, 
boxrule=0.8pt, width=\textwidth,
title=Prompt for Proxy Reward Evaluation
] 
"""\\
\textbf{\# Task: Evaluate Memory Potential}\\
You are an expert on human memory.
A system just observed this turn of dialogue:
---\\
\{history\_str\}\\
Turn to Evaluate: \{turn\_content\}\\
---\\

\textbf{\# Question}\\
How valuable is the "Turn to Evaluate" for saving to a long-term memory, to be used in *future* conversations?\\
Consider:\\
- High Value (4-5): Contains specific facts, preferences, plans, or significant emotional moments.\\
- Low Value (1-2): Simple acknowledgement ("OK", "I see"), filler, or common knowledge.\\

Rate this turn's memory potential on a scale of \{min\} to \{max\}. Your response MUST be only the number (e.g., "4.5").\\

Rating:\\
"""
\end{tcolorbox} 
\caption{Prompt for providing immediate proxy rewards to the \texttt{Planner} agent.}
\label{fig:Proxy Reward}
\end{figure*}

During online interaction, the \texttt{Planner} requires immediate feedback to guide its exploration of information value. We employ an LLM-based judge to provide a proxy reward ($r^{proxy}$), as specified in Figure \ref{fig:Proxy Reward}. This evaluation focuses on identifying specific facts or emotional markers that warrant long-term storage.

\subsection{\texttt{Trigger} Agent: Quality Reward Evaluation}
\begin{figure*}[!t]
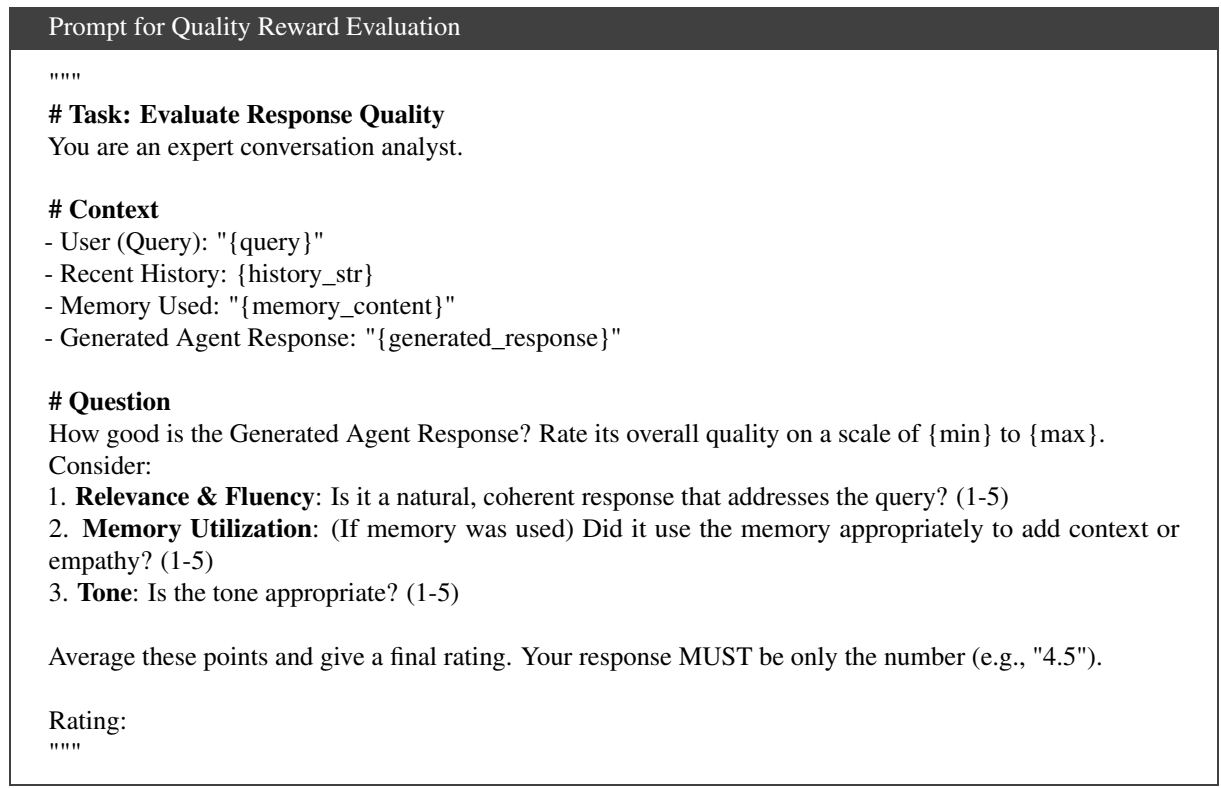

\begin{tcolorbox}[
colframe=black!75!white, 
colback=white, sharp corners, 
boxrule=0.8pt, width=\textwidth,
title=Prompt for Quality Reward Evaluation
] 
"""\\
\textbf{\# Task: Evaluate Response Quality}\\
You are an expert conversation analyst.\\

\textbf{\# Context}\\
- User (Query): "\{query\}"\\
- Recent History: \{history\_str\}\\
- Memory Used: "\{memory\_content\}"\\
- Generated Agent Response: "\{generated\_response\}"\\

\textbf{\# Question}\\
How good is the Generated Agent Response? Rate its overall quality on a scale of \{min\} to \{max\}.\\
Consider:\\
1. \textbf{Relevance \& Fluency}: Is it a natural, coherent response that addresses the query? (1-5)\\
2. \textbf{Memory Utilization}: (If memory was used) Did it use the memory appropriately to add context or empathy? (1-5)\\
3. \textbf{Tone}: Is the tone appropriate? (1-5)\\

Average these points and give a final rating. Your response MUST be only the number (e.g., "4.5").\\

Rating:\\
"""
\end{tcolorbox} 
\caption{Prompt for evaluating response quality and providing feedback to the \texttt{Trigger} agent.}
\label{fig:Quality Reward}
\end{figure*}

To align the \texttt{Trigger}'s retrieval policy with human preferences, the system evaluates the final response quality ($r^{qual}$). The prompt shown in Figure \ref{fig:Quality Reward} directs an LLM judge to score the response based on relevance, fluency, and the appropriate utilization of memory.

\end{document}